\documentclass{article}

 \usepackage[preprint]{neurips_2026}

\usepackage[utf8]{inputenc} 
\usepackage[T1]{fontenc}    
\usepackage{hyperref}       
\usepackage{url}            
\usepackage{booktabs}       
\usepackage{amsfonts}       
\usepackage{nicefrac}       
\usepackage{microtype}      
\usepackage{xcolor}         
\usepackage{microtype}
\usepackage{graphicx}
\usepackage{caption}
\usepackage{subfigure}
\usepackage{booktabs} 
\usepackage{listings}
\usepackage{multirow} 
\usepackage{colortbl}
\usepackage{relsize}   
\usepackage{amsmath}
\usepackage{siunitx}
\usepackage{wrapfig}
\usepackage[most]{tcolorbox}
\usepackage{fvextra}
\usepackage{caption}

\newtcolorbox{fileviewerbox}[1]{
  title={#1},
  colback=gray!5,
  colframe=gray!60,
  fonttitle=\ttfamily,
  breakable,
  enhanced
}

\newcommand{\decrease}[1]{\textcolor{red}{(-#1)}}
\newcommand{\smalltab}{\hspace{10pt}}

\title{AutoTrainess: Teaching Language Models to Improve Language Models Autonomously}

\author{Zhaojian Yu\thanks{~~Equal contribution.} , 
Penghao Yin$^*$,  Shuzheng Gao$^*$, Shilin He, Kai Cai,  Xiao-Ping Zhang\thanks{~~Corresponding author.}
\\
$^1$Tsinghua University
$^2$The Chinese University of Hong Kong
$^3$Simple Agent Lab
\\
  \texttt{zhaojian.yu01@gmail.com} \\
}

\begin{document}

\maketitle

\setcounter{footnote}{0}
\begin{abstract}
Training language models (LMs) remains a highly human-intensive process, even as frontier language model agents become increasingly capable at software engineering and other long-horizon tasks. A central challenge is that autonomous post-training is not just a coding problem: it requires the agent to repeatedly plan iterations, construct benchmark-aligned data, run stable training jobs, evaluate checkpoints, and preserve experiment state across many hours of interaction. We present AutoTrainess, a LM agent that exposes these operations as a repository of agent-computer interfaces for planning, data preparation, training, evaluation, and logging. 
Rather than leaving the agent to operate in a raw CLI environment with an underspecified action space, AutoTrainess externalizes prior human experience as explicit workflows, rules, and execution constraints that guide the agent toward effective and reliable training behavior.
On PostTrainBench, AutoTrainess consistently outperforms CLI-only baselines, achieving 26.94 average score with GPT-5.4 (Codex) versus 23.21 for CLI-only. It also generalizes across models and harnesses, improving DeepSeek-V4-Flash (OpenCode) from 12.13 to 19.58 \footnote{~~Code and data are avaliable at \url{https://github.com/simple-agent-lab/AutoTrainess}.}.

\end{abstract}

\begin{figure}[h]
    \centering
    \includegraphics[width=0.6\linewidth]{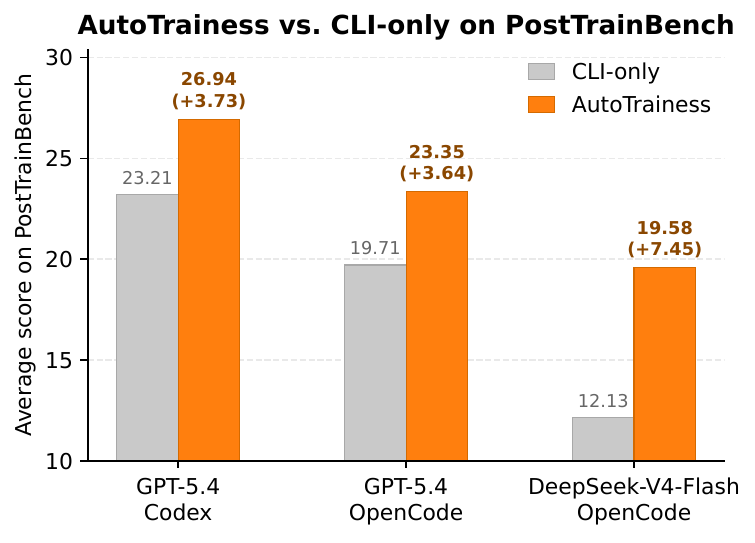}
    \caption{Comparison between AutoTrainess and CLI-only on PostTrainBench. Each agent is evaluated with  
    4 base models (Qwen3-1.7B, Qwen3-4B, SmolLM3-3B, and Gemma3-4B), access to an H20 GPU, and a 10-hour time limit to improve model performance through post-training. }
    \label{fig:teaser}
\end{figure}

\section{Introduction}
\begin{figure}[h]
    \centering
    \includegraphics[width=\linewidth]{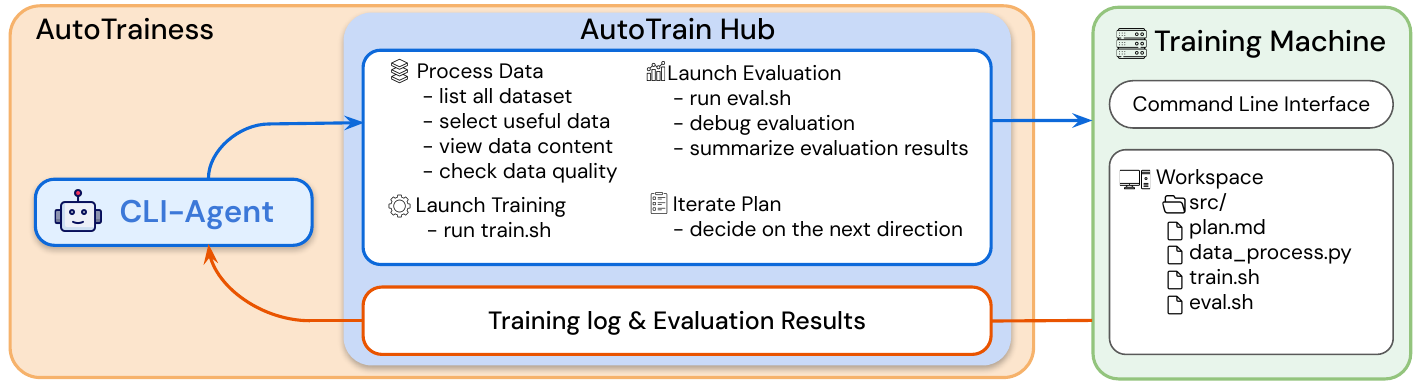}
    \caption{AutoTrainess is a LM agent that interacts with 
    training environments through a training-specialized Agent-Computer Interface, named AutoTrainHub. }
    \label{fig:teaser}
\end{figure}
Recent work has demonstrated the impressive capabilities of frontier LM agents in real-world automation tasks such as software engineering \cite{JimenezYWYPPN24,HuangEtAl2024MLAgentBench} or scientific discovery \cite{NovikovEtAl2025AlphaEvolve,YuEtAl2025AlphaResearch}. 
Despite this progress, there remains a persistent gap: as language models become increasingly powerful, the process of improving them still relies on extensive human effort. 
One possible pathway for LMs' self-improvement is to treat LM training as a software engineering task, leveraging coding agents to autonomously generate and optimize training code.
However, successful LLM training requires more than strong coding ability: experienced human engineers also rely on substantial accumulated training expertise, empirical intuitions and artifacts (e.g., sophisticated data generation pipelines).
Inspired by this observation, we investigate whether LM agents can similarly benefit from prior experience and artifacts created by human researchers when performing autonomous training tasks.

Consider the simple setting of an agent interacting directly with a command-line interface (CLI) on Linux. We find that even powerful agents struggle to 
manage training tasks effectively. For example, when creating custom training datasets, the agent may introduce errors in packing sequences to maximum length or fail to use correct 
chat templates, resulting in suboptimal training data and frequent dataloader exceptions. 
These limitations suggest that model self-improvement remains difficult without access to human expertise,
highlighting the necessity of a training-aware agent-computer interface (ACI) \cite{yang2024sweagent} that packages human-curated best practices into agent harness. 

We introduce AutoTrainess, an LM agent empowered by a training-specialized ACIs that enables autonomous end-to-end LLM post-training. 
In contrast to CLI-only agent's highly uncertain action space,
AutoTrainess’s ACI provides a set of semantically meaningful training heuristics 
supported by standard pipelines for data preparation, model training, and evaluation.
Through AutoTrainess's rich human prior knowledge on training dynamics and data processing, the agent can diagnose issues and refine training strategies efficiently, enabling stable and iterative model improvement with reduced human supervision.

Using GPT-5.4 (Codex) as backbone, AutoTrainess achieves 26.94 average on PostTrainBench \cite{RankEtAl2026PostTrainBench}, outperforming the CLI-only baseline (23.21). We conduct ablation studies on PostTrainBench Qwen3-4B subset and confirm that the ACIs contributes 3–8 point improvements. To validate the generalization of our ACIs, 
we show that it is portable to a different LM; AutoTrainess powered by DeepSeek-V4-Flash (OpenCode) still attains 19.58 on average, significantly outperforming the CLI-only baseline of 12.13 while maintaining stable end-to-end training with minimal human oversight.

Our contributions are twofold. First, we identify the fundamental limitations of raw CLI interfaces in LLM training tasks and introduce a training-specialized Agent-Computer Interface (ACI) that embeds rich human expertise directly into the agent-environment interaction. Second, we present AutoTrainess, which achieves strong results on PostTrainBench, enables stable and reliable autonomous end-to-end post-training, and provides a practical pathway toward scalable LLM self-improvement.







\section{AutoTrainess}
We investigate whether autonomous LLM training can benefit from externally provided prior experience, much as human engineers draw upon accumulated workflows and artifacts rather than constructing each training pipeline from scratch.

To this end, we instantiate human experience as a repository of reusable ACIs, named AutoTrainHub, that provide the agent with structured guidance for conducting iterative post-training. 
Rather than leaving the agent to solve training as
an unconstrained coding task, AutoTrainHub expose a stage-wise interface over the workspace, specifying what
artifacts should be used, what outputs should be produced, and what operational constraints must be respected.
AutoTrainHub organizes autonomous training into a closed-loop workflow with four modules: \texttt{data process}, \texttt{training}, \texttt{evaluation}, and \texttt{logging\&planning}. At a high level, the agent first analyzes evidence from prior experiments to define the goal of the next iteration, then prepares
training data, runs model training, evaluates the resulting checkpoint on the benchmark's real evaluation pipeline, and finally records the completed iteration in a structured experiment log. 

\textbf{Data processing.}
We introduce three explicit actions to construct task-aligned training dataset : data selection, data construction, and data validation.
The data selection action identifies the problems or behaviors implied by previous failures and selects initial source
directions for data construction, such as existing local data, externally collected data, or model-distilled data.
The data construction action supports a bounded set of dataset operations, including extraction, cleaning,
deduplication, rewriting, restructuring, synthesis, distillation, and schema normalization. Before deciding the final
training sample format, the agent must inspect the benchmark's actual evaluation interface, such as evaluation scripts,
chat templates, or task-context files.
The data validation action first checks whether the constructed examples match the benchmark-facing task interface and
rendered examples. It then filters out low-quality or risky samples, including garbage text, corrupted examples,
duplicates, unrealistic synthetic patterns, and potential data leakage.
The validation stage returns one of three outcomes: approval for training, return to construction, or return to
selection. This explicit return mechanism allows the agent to distinguish between execution errors within a viable data direction and failures caused by an incorrect data optimization direction itself.

\textbf{Training.}
The training interface provides a stable training entry based on LlamaFactory \cite{ZhengZZYL24}. 
Rather than allowing the agent to freely choose training frameworks or implement custom loops, the interface fixes LlamaFactory as the training backend and provides dedicated scripts for installation and execution. This reduces engineering variance and makes autonomous training more reproducible.
 For supervised fine-tuning, the interface requires full-parameter fine-tuning, a small validation run before scaling up, and export of an evaluation-ready final model. For reinforcement learning, this interface is only applicable when supported by recent evaluation evidence. In such cases, the agent must explicitly specify the reward definition or feedback signal actually used during the run.
In both cases, failures must be debugged within the same LlamaFactory-based workflow rather than bypassed by switching to another framework. This encodes a practical prior commonly used by human practitioners: in iterative post-training, maintaining a stable and comparable training results is often more valuable than maximizing implementation flexibility.

\textbf{Evaluation.}
The evaluation interface runs the trained checkpoint on the benchmark's real evaluation pipeline and records the evidence needed for the next iteration. Its primary role is to ensure that downstream decisions are based on comparable and sufficiently informative evaluation results.
Specifically, the interface requires the agent to evaluate the final model using the benchmark's canonical entrypoint, save raw outputs  under an evaluation results directory, and produce a concise evaluation summary. 
In addition, the agent must generate a compact summary containing 15 randomly selected evaluation examples with score, input, target, and model output. 
When compatible evaluation logs are available, a log parser script is used to extract these samples into an inspection artifact. 
Finally, the agent is required to summarize the main observed failure modes and classify each as primarily a data problem, a training problem, or an inference or template problem. This structured diagnostic output directly informs the next planning and new decisions.

\textbf{Logging \& Planning.}
The logging interface appends one structured entry to the experiment log after each completed iteration. Each entry records the iteration context, motivation, references consulted, starting checkpoint, training data, method, training configuration, evaluation protocol, result, analysis, generated artifacts, and next action.
This persistent log serves as a compact long-horizon memory over the training process. It preserves concrete evidence across iterations, supports reproducibility and retrospective analysis, and provides subsequent agent runs with a structured summary of prior decisions and their outcomes.
The planning interface defines the objective of the next experiment iteration based on empirical evidence from prior runs. Given previous evaluation results, training outcomes, and the current workspace state, the agent is required to identify the main observed problems, decide the primary objective of the current iteration, specify the planned intervention, and define a concrete success criterion.


AutoTrainHub treats human experience not as latent capability inside the base model, but as an explicit external scaffold for autonomous training.
It integrates researcher experience about how to plan iterations, construct benchmark-aligned data, run stable training, perform evidence-grounded evaluation, and preserve experimental memory. 
By exposing these prior human experience through explicit ACIs, we transform autonomous LLM training from an open-ended software engineering problem into a structured sequential decision process. 
This setup allows us to directly study whether such externalized prior experience improves the effectiveness and efficiency of agentic self-improvement.

\section{Experiments}

\begin{table}[t]
\caption{Main results on PostTrainBench. Each agent receives
4 base models and 7 benchmarks for each model, access to an H20 GPU, and a 10-hour time limit. Detailed Results are shown at \autoref{app-detail-res}.}
\centering
\label{tab:main}

\resizebox{\linewidth}{!}{%
\begin{tabular}{lccccc}
\toprule
 & \multicolumn{5}{c}{\textbf{PostTrainBench}} \\
\cmidrule(lr){2-6}
\textbf{Harness} & Qwen3-1.7B & Qwen3-4B & SmolLM-3B & Gemma-4B & \textbf{Avg(\%).} \\
\cmidrule(lr){1-1} \cmidrule(lr){2-6}

Instruct     & 49.41 & 63.75 & 44.81 & 46.58 & 51.14 \\
Base    & 6.66  & 14.34 & 4.52  & 4.60  & 7.53 \\
\cmidrule(lr){1-1} 
CLI-only    &   &   &   &   &   \\
\smalltab w/ GPT-5.4 (Codex)   & 16.90 & 27.09 & 23.96 & 24.88 & 23.21 \\
\smalltab w/ GPT-5.4 (OpenCode)   & 20.01    & 17.01    & 19.51    & 22.32    & 19.71 \\
\smalltab w/ DeepSeek-V4-Flash (OpenCode) & 8.14    & 15.18    &  14.77   & 10.43   & 12.13 \\
\cmidrule(lr){1-1} 
AutoTrainess &     &     &     &     &  \\
\smalltab w/ GPT-5.4 (Codex) & 25.67 & 32.60 & 25.60 & 23.88 & 26.94 \\
\smalltab w/ GPT-5.4 (OpenCode)   & 22.08    & 25.91    & 24.20    & 21.20    & 23.35 \\
\smalltab w/ DeepSeek-V4-Flash (OpenCode) & 16.72    & 21.76    & 15.82    & 24.01    & 19.58 \\
\bottomrule
\end{tabular}
}
\end{table}

\subsection{Evaluation Setup}
\textbf{Datasets.} We primarily evaluate on PostTrainBench \cite{RankEtAl2026PostTrainBench}, which pairs each run with one of seven target benchmarks: AIME 2025~\cite{RankEtAl2026PostTrainBench}, ArenaHard~\cite{LiEtAl2025ArenaHard}, BFCL~\cite{PatilEtAl2025BFCL}, GPQA~\cite{ReinEtAl2023GPQA}, GSM8K~\cite{CobbeEtAl2021GSM8K}, HealthBench~\cite{AroraEtAl2025HealthBench}, and HumanEval~\cite{ChenEtAl2021HumanEval}. These benchmarks cover mathematical reasoning, function calling, general knowledge, health, and code generation. 

\textbf{Models.} PostTrainBench \cite{RankEtAl2026PostTrainBench} evaluates post-training over four base models: Qwen3-1.7B, Qwen3-4B, SmolLM3-3B, and Gemma-3-4B. All results, ablations, and analyses in this paper are based on two leading CLI agent scaffolds, Codex and OpenCode. We experimented with a number of closed and open-source agent backbones, including GPT-5.4 and DeepSeek-V4-Flash. We also tried smaller agent models such as Qwen3.5-35B-A3B, but found their performance in this autonomous training setting to be subpar. 

\textbf{Baselines.} We compare AutoTrainess to CLI-only harness, where agents can only use the internal tools and system prompt to interact with the autonomous training environment. The CLI-only harness system asks the LM to launch a training process by writing and run code with a shell process in sandbox.

\textbf{Metrics.}We report the aggregated scores of PostTrainBench. 
First, for each agent, we average its performance on each benchmark across the four base models (Qwen3-1.7B, Qwen3-4B, SmolLM3-3B, and Gemma-3-4B), yielding the per-benchmark agent score \( s_i^{\rm agent} \).
Next, we compute the benchmark weights $w_i$ as follows:
\[
w_i = \frac{1}{s_{\rm instruct}^i - s_{\rm base}^i}, \quad
\hat{w}_i = \frac{w_i}{\sum_j w_j}
\]
where \( s_{\rm instruct}^i \) and \( s_{\rm base}^i \) are the scores of the official instruction-tuned model and the base model on benchmark \( i \), respectively. This weighting scheme assigns higher weights to harder benchmarks where instruction tuning yields smaller gains.
The final aggregated score for each agent is then defined as:
\[
\text{Score}_{\rm agent} = \sum_i \hat{w}_i \cdot s_i^{\rm agent}
\]
This score provides a comprehensive measure of the agent's post-training effectiveness under the GPU constraint.

\subsection{Main Results}
Across all systems, AutoTrainess w/ GPT-5.4 (Codex) achieves the best performance all-around, getting 26.94 overall score of the PostTrainBench full set and 32.60 of the Qwen3-4B subset. As shown in \autoref{tab:main}, compared to CLI-only, AutoTrainess yields 15\% improved overall score relatively.  An LM-friendly Training ACI is confirmed by AutoTrainess's 15\%-25\% relative increase compared to CLI-only, with small decrease (< 5\%) on Gemma-4B subset. On the OpenCode scaffold, AutoTrainess shows consistent performance gains, highlighting its robust generalization capability across various CLI agents.



\subsection{Ablations}

We perform several ablations of the AutoTrainess interfaces, specifically with respect to the AutoTrainess w/ GPT-5.4 (Codex) configuration, summarized in \autoref{tab:combined_ablations_summary}. Our further analysis shed light on interesting agent behavior along with the impact of different harness designs.
\begin{wraptable}{r}{0.48\textwidth}  
\centering
\caption{PostTrainBench (Qwen3-4B) performance under ablations to the interfaces in AutoTrainess w/ GPT5.4 (Codex).}
\label{tab:combined_ablations_summary}

\small
\begin{tabular}{ll}
\toprule
\textbf{Method} & \textbf{Overall}  \\
\midrule
CLI-only             & 26.7 \\
AutoTrainess          & 32.6  \\
\quad w/o \texttt{data processing}   & 29.1 \footnotesize{\decrease{3.5}} \\
\quad w/o \texttt{training}  & 20.2 \footnotesize{\decrease{12.4}}  \\
\quad w/o \texttt{evaluation}   & 24.0 \footnotesize{\decrease{8.6}}  \\
\quad w/o \texttt{logging\&planning}  & 24.1 \footnotesize{\decrease{8.5}} \\

\bottomrule
\end{tabular}
\end{wraptable}

\begin{figure}[t]
    \centering
    \includegraphics[width=0.49\linewidth]{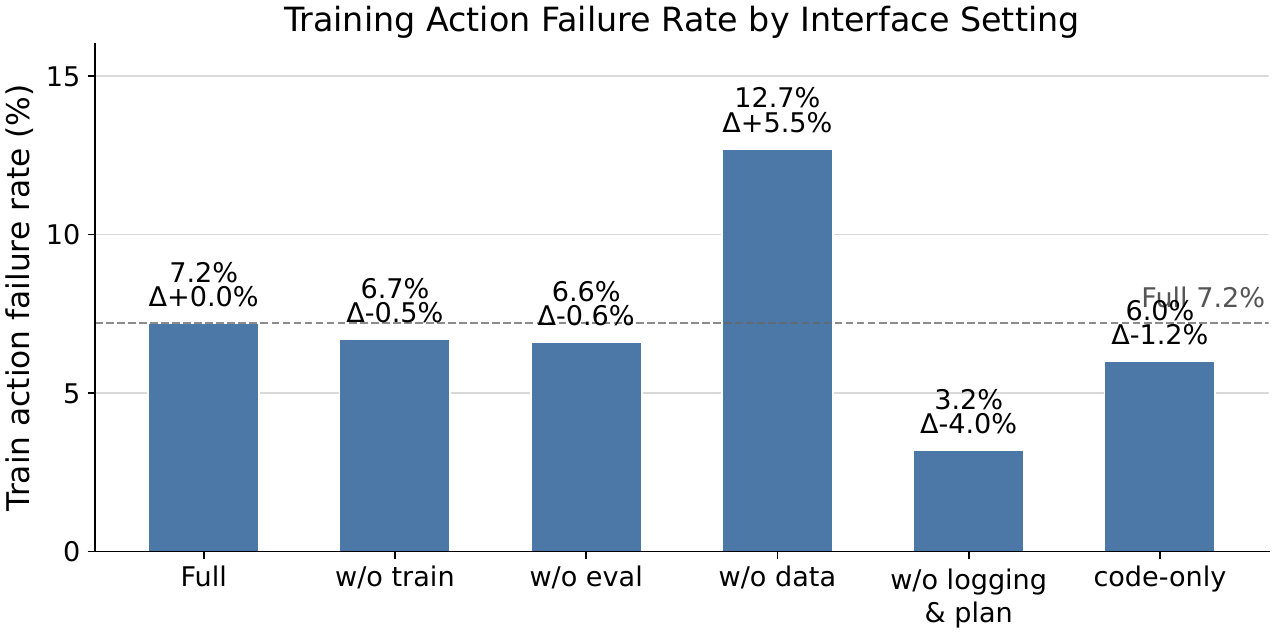}
    \includegraphics[width=0.49\linewidth]{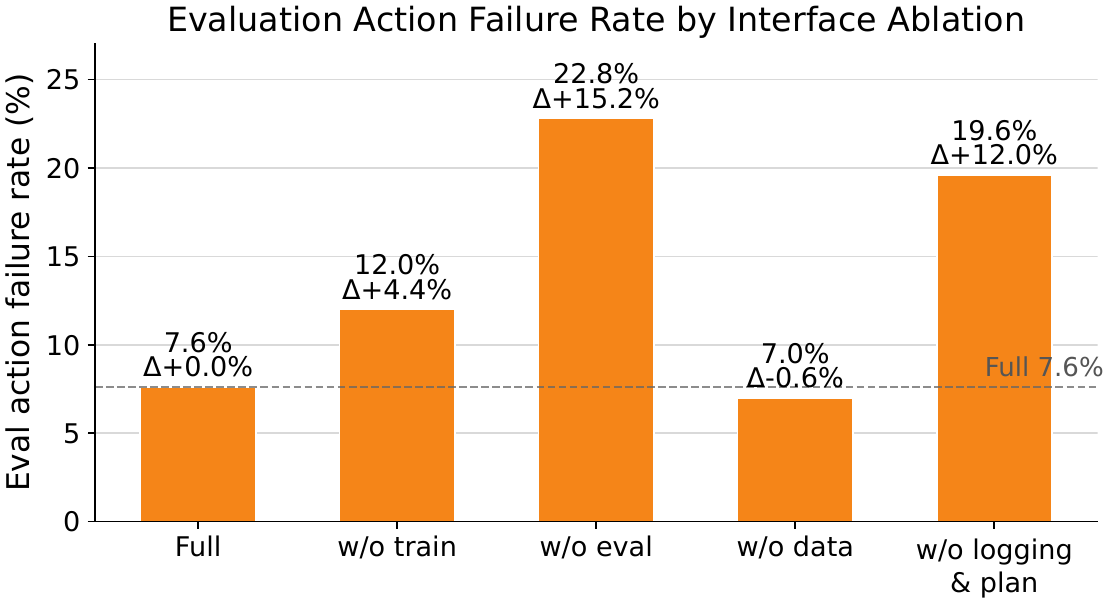}

    \caption{
    Failure rate under interface ablations.
    Each bar reports the action failure rate, and the second percentage is the difference from the full interfaces.
    }
    \label{fig:interface-train-action-failure-rate}
\end{figure}

\textbf{Data interfaces mainly protect the training input contract.}
Removing the \texttt{data} interface produces the largest increase in train action failure rate, from \num{7.2}\% to \num{12.7}\%, a difference of \num{5.5} percentage points.
At the same time, the evaluation action failure rate remains close to the full interface (\num{7.0}\%, \num{-0.6} percentage points from full).
As shown in \autoref{tab:data-action-counts}, the data interface leads the agent to engage substantially more in data-centric operations prior to training, including dataset inspection, cleaning, preference-pair construction, and synthetic data generation. This behavior suggests that the agent reasons more explicitly about dataset content, format, metadata, and sampling strategies before passing the data to the training pipeline.
Without this interface, errors about data loading are more likely to be deferred until debug or validation training, where they appear as training action failures rather than evaluation action failures.

\textbf{Training interfaces have weaker direct effects on train failures but affect downstream evaluation.}
The w/o \texttt{train} interface has a train action failure rate similar to the full interface (\num{6.7}\%, \num{-0.5} percentage points from full), suggesting that this interface does not strongly reduce per-command training action failures in the aggregate.
However, its evaluation action failure rate rises to \num{12.0}\%, a difference of \num{4.4} percentage points from full.
This suggests an indirect role: the \texttt{train} interface helps standardize checkpoint locations, merged-model outputs, and final-model handoff conventions.
When it is removed, training commands do not necessarily fail more often, but their artifacts can become more ambiguous or inconsistently referenced by downstream evaluation commands.
Overall, the ablation results suggest that each interface is most effective at guarding its corresponding boundary: the \texttt{eval} and \texttt{logging\&plan} interfaces reduce evaluation action failures, the \texttt{data} interface reduces training-input action failures, and the \texttt{train} interface stabilizes the artifact handoff into evaluation.

\textbf{Evaluation interfaces provide the strongest protection against evaluation action failures.}
Figure~\ref{fig:interface-train-action-failure-rate} shows that removing the \texttt{eval} interface causes the largest increase in evaluation action failures.
The full interface has an evaluation action failure rate of \num{7.6}\%, while the w/o \texttt{eval} interface increases this rate to \num{22.8}\%, a difference of \num{15.2} percentage points.
By contrast, the corresponding train action failure rate remains close to the full interface (\num{6.6}\% vs. \num{7.2}\%; Figure~\ref{fig:interface-train-action-failure-rate}).
This asymmetry indicates that the \texttt{eval} interface primarily improves the reliability of evaluation orchestration rather than general command execution.
Its contribution is to preserve benchmark-specific invocation conventions, limited-evaluation checks, output-path discipline, and model-serving prerequisites that are easy to violate when evaluation commands are assembled without a dedicated interface.

\textbf{Logging and planning interfaces stabilize stateful evaluation loops.}
The w/o \texttt{logging\&plan} interface also substantially increases the evaluation action failure rate, from \num{7.6}\% to \num{19.6}\%, a difference of \num{12.0} percentage points.
Unlike the w/o \texttt{eval} setting, this ablation does not remove the ability to call evaluation scripts.
Instead, it weakens the agent's ability to maintain state across iterations: which model artifact should be evaluated, whether an output directory has been created, which vLLM server is active, and which previous fixes should be reused.
This makes evaluation attempts more brittle, especially for full and limited evaluations that depend on a consistent train-to-eval handoff.
The lower train action failure rate in this condition (\num{3.2}\%, \num{-4.0} percentage points from full) should not be interpreted as improved training reliability, since this interface ablation also runs fewer train and evaluation commands.




\subsection{Exploration vs. Exploitation}
\begin{figure}[t]
    \centering
    \begin{minipage}{0.48\textwidth}
        \centering
        \caption{Ablation on data related action counts in PostTrainBench trajectories. }
        \label{tab:data-action-counts}
        \begin{tabular}{lcc}
        \toprule
          & \multicolumn{2}{c}{\textbf{AutoTrainess}} \\
        \textbf{Action} &  full &  w/o data \\
        \midrule
        Read dataset content & 725 & 579  \footnotesize{\decrease{20.1\%}} \\
        Clean dataset & 95 & 71 \footnotesize{\decrease{25.3\%}} \\
        Build preference pairs & 44 & 32 \footnotesize{\decrease{27.3\%}}  \\
        Synthesize Data & 22 & 4 \footnotesize{\decrease{81.8\%}} \\
        Find dataset sources & 18 & 17 \footnotesize{\decrease{5.6\%}}\\
        Add corrective data & 18 & 16 \footnotesize{\decrease{11.1\%}} \\
        \bottomrule
        \end{tabular}
    \end{minipage}
    \hfill
    \begin{minipage}{0.48\textwidth}
        \centering
        \includegraphics[width=\linewidth]{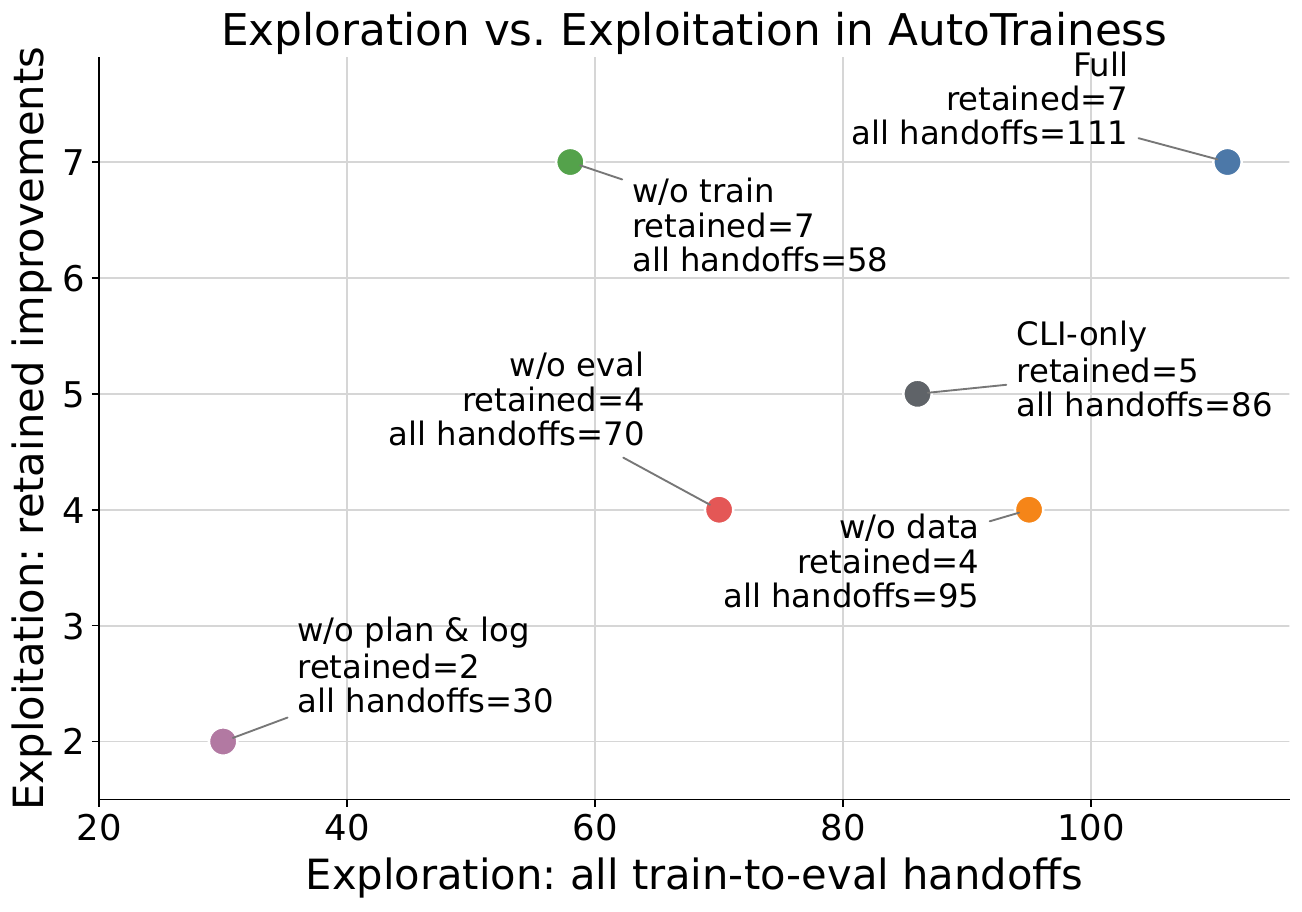}
        \caption{Exploration exploitation balance under interface ablations.
        Exploration is measured by the number of train-to-eval handoffs, while exploitation is measured by retained improvements.}
        \label{fig:codex-explore-exploit-balance}
    \end{minipage}
    
\end{figure}

As shown in Figure~\ref{fig:codex-explore-exploit-balance}, the full interface achieves the highest exploration (111 train-to-eval handoffs) while retaining 7 improvements, corresponding to a 6.3\% retained-improvement yield. This demonstrates that the complete interface set enables broad search without sacrificing the ability to convert promising trials into durable gains.Removing the data interface maintains relatively high exploration (95 handoffs) but reduces retained improvements to 4 (4.2\% yield). This suggests the data tools primarily enhance the quality of explored configurations by supporting standardized dataset construction, cleaning, and formatting, rather than merely increasing trial volume.In contrast, ablating the train interface sharply reduces exploration (58 handoffs) while preserving all 7 improvements, raising the yield to 12.1\%. The training interface thus primarily broadens the search frontier by facilitating more training configurations and checkpoint variants.The eval interface plays a key role in selection: its removal lowers both handoffs (70) and retained improvements (4). Similarly, disabling logging\&plan causes the most severe collapse (30 handoffs, 2 improvements), underscoring its importance for maintaining iterative state and closed-loop coordination.The CLI-only baseline (86 handoffs, 5 improvements) offers moderate performance but falls short of the full interface on both dimensions. Overall, the ablation results reveal complementary contributions: the train interface drives exploration breadth, data and eval tools improve the conversion of trials into retained gains, and logging \& plan sustains the iterative search process.

\begin{figure}[t]
    \centering
    \includegraphics[width=1\linewidth]{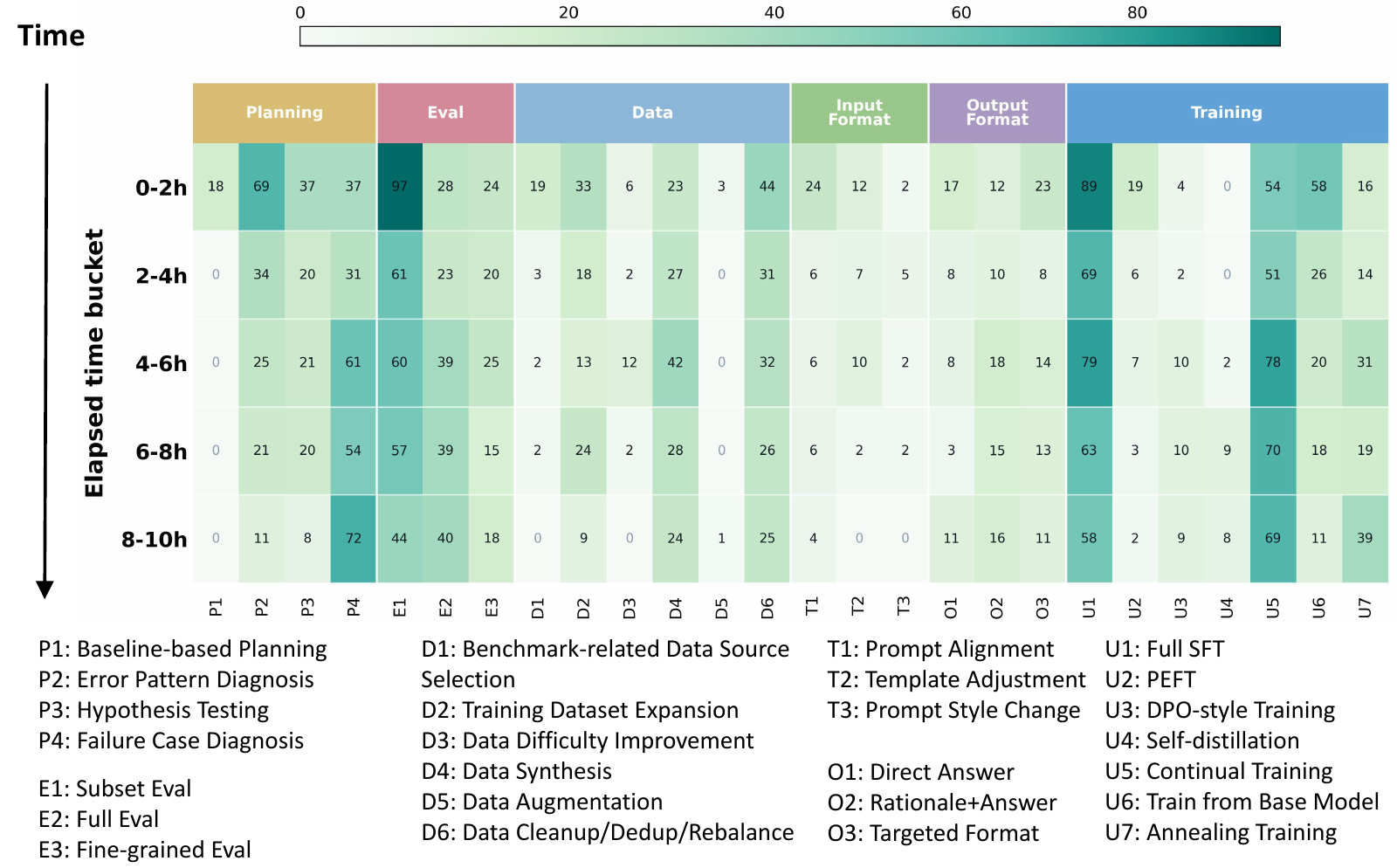}
    \caption{Frequency of different agent behaviors in AutoTrainess at different time stages.}
    \label{fig:behavior}
\end{figure}

\subsection{Analysis of Agent Behavior}

\begin{figure}[t]
    \centering
    \subfigure[Qwen3-4B on GPQA]{\includegraphics[width=0.49\linewidth]{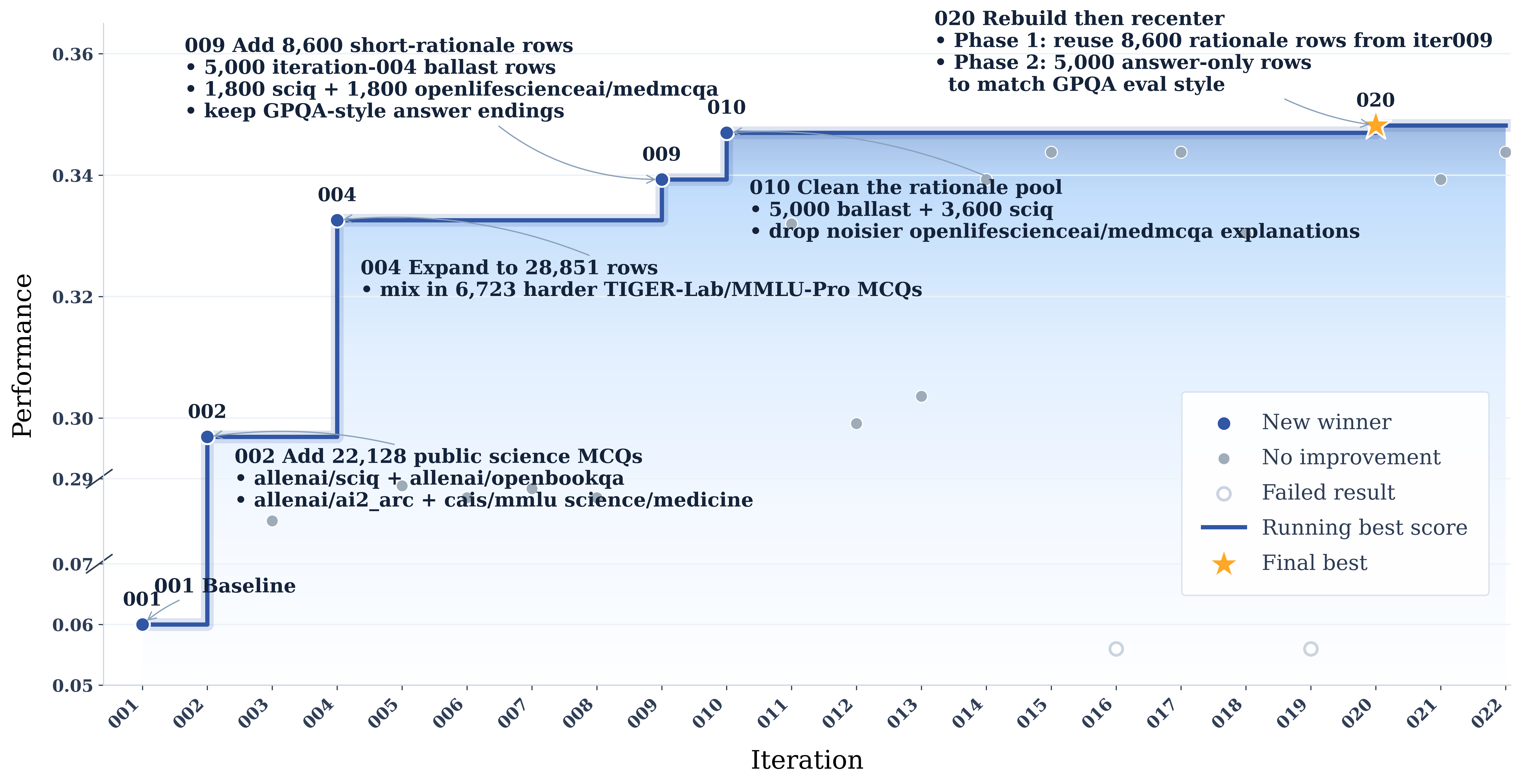}}
    \hfill
    \subfigure[Qwen3-4B on GSM8K]{\includegraphics[width=0.49\linewidth]{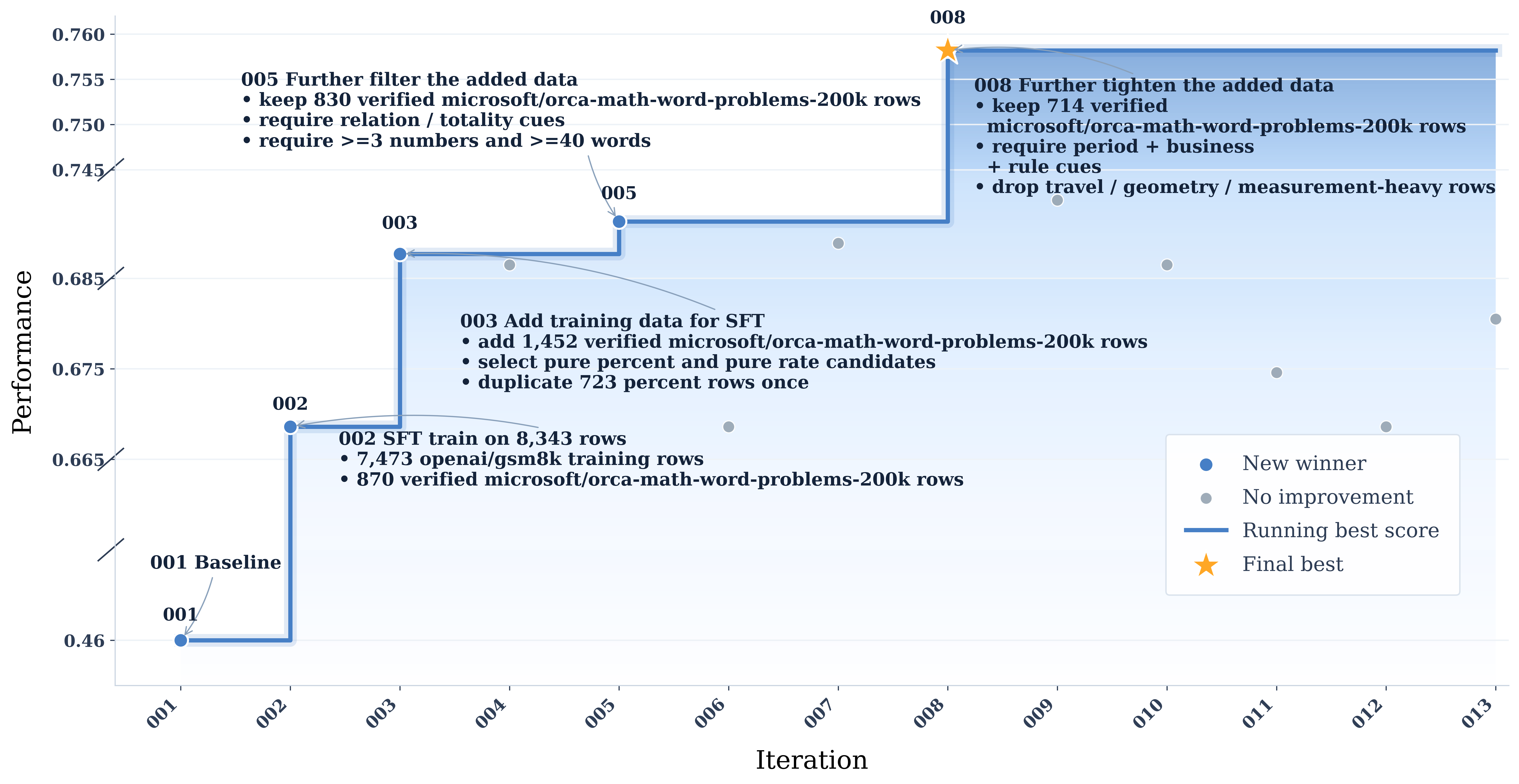}}

    \vspace{0.5em}

    \subfigure[Gemma-4B on HumanEval]{\includegraphics[width=0.49\linewidth]{figures/performance_iteration_humaneval.png}}
    \hfill
    \subfigure[Gemma-4B on ArenaHard]{\includegraphics[width=0.49\linewidth]{figures/performance_iteration_arenahard.png}}
    \caption{Examples of the model training process of AutoTrainess across different datasets.}
    \label{fig:intro-performance}
\end{figure}

We further analyze agent behavior under AutoTrainess using training trajectories produced by GPT-5.4 with Codex. From the concrete actions in these trajectories, we abstract a higher-level taxonomy with 6 major classes and 26 subclasses. We then use an LLM-as-a-judge to annotate the behavior exhibited at each iteration across different model--dataset training runs. Finally, we group iterations by their position in the training timeline to obtain the stage-wise behavior distribution shown in Figure~\ref{fig:behavior}. Based on this, we achieve the following findings:

\textbf{Agents first adapt to the benchmark before pursuing optimization.}
At the beginning of a run, agents primarily calibrate the task interface. In the first two hours, baseline-based planning ($P1$) appears $18$ times, but disappears entirely in the remaining four blocks. Benchmark-interface work follows the same front-loaded pattern. Behaviors like Benchmark prompt alignment ($T1$), Template change ($T2$) and Benchmark-near data selection ($D1$) are more frequently distributed at the beginning, with their frequencies dropping rapidly thereafter. For tasks with time-consuming evaluations, agents often rely on lightweight validation ($E1$) over a small subset, allowing them to obtain feedback quickly and iterate within the limited time budget. These patterns indicate that agents typically begin by running the base model, aligning prompts and templates with the benchmark, and setting up a lightweight validation loop before attempting more specialized interventions. For example, in the GPQA run shown in Figure~\ref{fig:intro-performance}, the early gains come from adding benchmark-aligned science multiple-choice data and preserving GPQA-style answer formatting.

\textbf{Mid-run strategies shift toward targeted data construction and training.}
After the training format stabilizes and the agent obtains error signal, agent's strategies move toward data construction and optimization. For example, the data synthesis ($D4$), which rises from $19$ occurrences in first two hours to $40$ in the next two hours and remains high at $36$ in the next time block. Agents also start to explore a broader range of training strategies during this phase. DPO-style training ($U3$) \cite{RafailovSMMEF23} is absent in the first two hours, and then appears $7$ and $12$ times in block 2 and 3, respectively. Self-distillation updates \(U4\) 
exhibit a similar trend, suggesting that agents increasingly experiment with more targeted optimization methods once they have identified actionable weaknesses. This pattern is also visible in Figure~\ref{fig:intro-performance}: in the HumanEval example, the agent first normalizes the task format and then improves further by adding benchmark-guided and synthetic programming tasks.

\textbf{Later iterations concentrate on remaining failures and targeted fixes.}
Late iterations focus on improving the current best model by analyzing its remaining failure cases and adding targeted training data. For example, failure case diagnosis ($P4$) that analyzes the left error samples, increases almost monotonically. full-benchmark evaluation ($E2$) also becomes more prominent later in the run. This indicates that agents are more inclined to conduct full-scale evaluations near the end of a run, in order to obtain a more comprehensive understanding of the model's shortcomings and avoid being misled by small validation sets. For example, in the ArenaHard trajectory in Figure~\ref{fig:intro-performance}, later gains come from narrowing the data mixture toward rewrite and style-control examples and then switching from full fine-tuning to LoRA to preserve base writing quality.

\begin{figure}[t]
    \centering
    \includegraphics[width=1\linewidth]{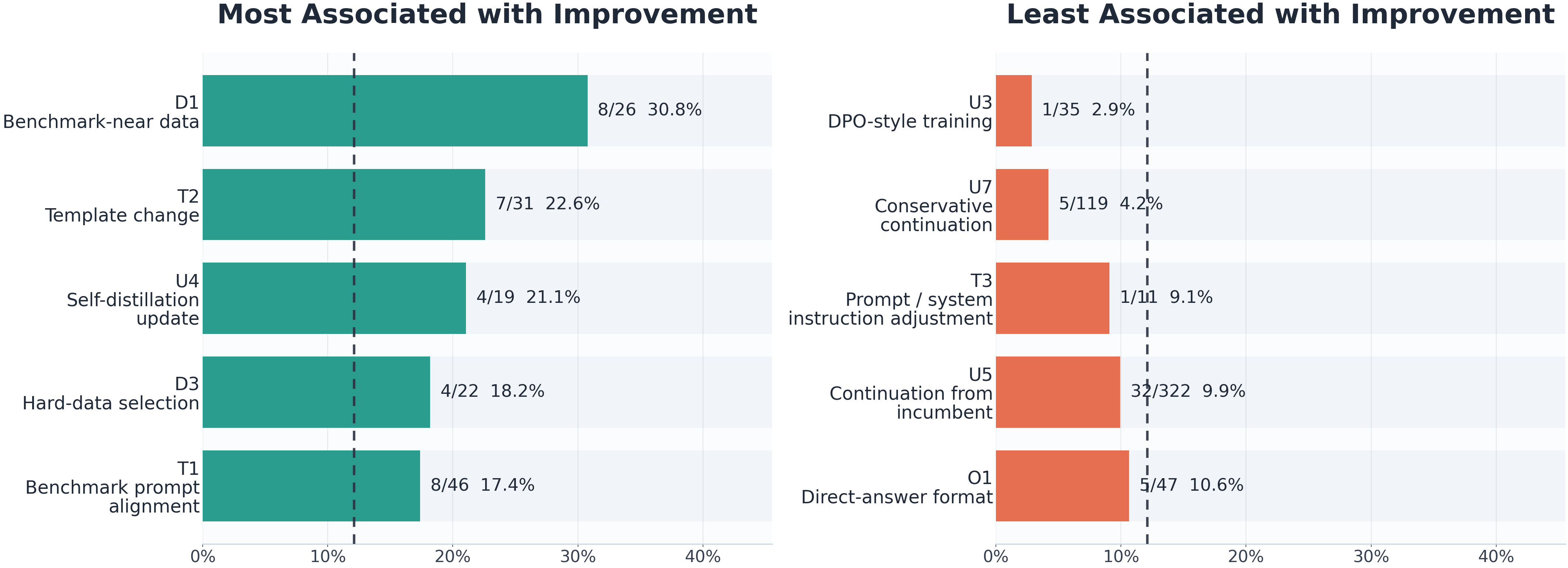}
    \caption{Statistics of agent behaviors most and least correlated with performance improvement.}
    \label{fig:good-bad}
\end{figure}

\textbf{Performance gains are primarily driven by benchmark alignment and targeted correction.}
Figure~\ref{fig:good-bad} further breaks down which behaviors are more often associated with improvements versus regressions. Among non-planning and non-evaluation strategies, the behaviors most strongly associated with improvements are benchmark-related data ($D1$, $8/26$ improving occurrences), template change ($T2$, $7/31$), self-distillation update ($U4$, $4/19$), data difficulty improvement ($D3$, $4/22$), and benchmark prompt alignment ($T1$, $8/46$). These strategies reduce mismatch between training and evaluation or directly address previously identified weaknesses. Instead, DPO-style training ($U3$) appears in only $1/35$ improving occurrences, making it one of the weakest strategies by this criterion. Annealing training ($U7$) also yields limited gains, with only 5 improving occurrences out of 119, suggesting that agents may still struggle to configure effective annealing data and training hyperparameters.

\textbf{Agents exhibit training habits distinct from human workflows.}
Agents show a strong preference for continuing from the current best checkpoint instead of rebuilding the data pipeline and restarting from the base model. Specifically, continual training ($U5$) from the best model appears $322$ times in the pooled trajectories, whereas retraining from base ($U6$) appears only $133$ times. This suggests that, once a promising checkpoint is found, the agent usually treats it as the anchor for subsequent exploration rather than repeatedly resetting optimization from scratch. This preference is likely tied to the limited time budget of the setting: with only around ten hours available, continuing from the strongest existing checkpoint is a much cheaper way to test new ideas than rerunning a full base-model training path after every data modification. Moreover, agents rarely use data augmentation in the conventional sense. Specifically, explicit augmentation or rewriting behavior ($D5$) appears only $4$ times in total. This is noticeably different from common human practice, where augmentation is often a standard method for improving robustness or generalization of model.

\section{Related Work}

\paragraph{Automatic research agents.}
Beyond software engineering~\cite{JimenezYWYPPN24,XiaEtAl2024Agentless,GaoEtAl2024SearchBasedCodeOpt} or computer use~\cite{XieEtAl2024OSWorld,YangEtAl2025RiOSWorld}, recent work has begun to study whether LLM agents can automate research with limited human intervention. The AI Scientist casts scientific discovery as an open-ended loop that proposes ideas, runs experiments, analyzes outcomes, and drafts papers automatically \cite{LuLangeFoersterCluneHa2024}. OpenResearcher presents a research assistant for long-horizon literature exploration, evidence collection, and report generation \cite{ZhengEtAl2024OpenResearcher}, while a later work focuses on synthesizing long-horizon deep-research trajectories for training and evaluation \cite{LiEtAl2026OpenResearcherPipeline}. Gottweis et al. present AI co-scientist, a system that supports scientific hypothesis generation and iterative refinement in domain research workflows \cite{GottweisEtAl2025AICoScientist}. These works suggest that automated research is a long-horizon, knowledge-intensive problem in which agents must combine domain expertise with sustained coordination across literature understanding, experimentation, and iterative refinement.

\paragraph{Benchmarks for research and experimentation agents.}
Recently, there has been increasing interest in evaluating whether language agents can conduct long-horizon research work. For example, MLAgentBench investigates the ability of language agents to conduct machine learning experimentation \cite{HuangEtAl2024MLAgentBench}. CORE-Bench \cite{SiegelEtAl2024CoreBench} assesses whether agents can reproduce results from existing research artifacts. PostTrainBench further extends this line of evaluation to end-to-end LLM post-training workflows \cite{RankEtAl2026PostTrainBench}. Beyond these benchmarks, other recent efforts have broadened the evaluation landscape to cover a wider range of scientific and engineering activities, including machine learning engineering, scientific law discovery, and literature discovery \cite{ChanEtAl2025MLEBench,StaraceEtAl2025PaperBench,ZhengEtAl2025NewtonBench,XiongEtAl2026AutoResearchBench}. 

\section{Conclusion}

We presented AutoTrainess, a training-specialized agent-computer interface for autonomous LLM post-training. By externalizing prior human experience as reusable skills for planning\&logging, data preparation, training, and evaluation. AutoTrainess turns post-training into a more structured and reliable long-horizon workflow. Experiments on PostTrainBench show that AutoTrainess consistently outperforms CLI-only baselines, while ablations further indicate that data, training, evaluation, and planning interfaces each contribute meaningfully to the overall gains. 

\bibliographystyle{unsrt}
\bibliography{references}


\clearpage
\appendix

\section{Case Study on the Effects of Different Skills}

We use two representative examples to illustrate the role of different skills in AutoTrainess. The first compares the AutoTrainess against the data-skill ablation on ArenaHard, and the second compares the AutoTrainess against the eval-skill ablation on HealthBench. 

\begin{figure}[h]
    \centering
    \includegraphics[width=1\linewidth]{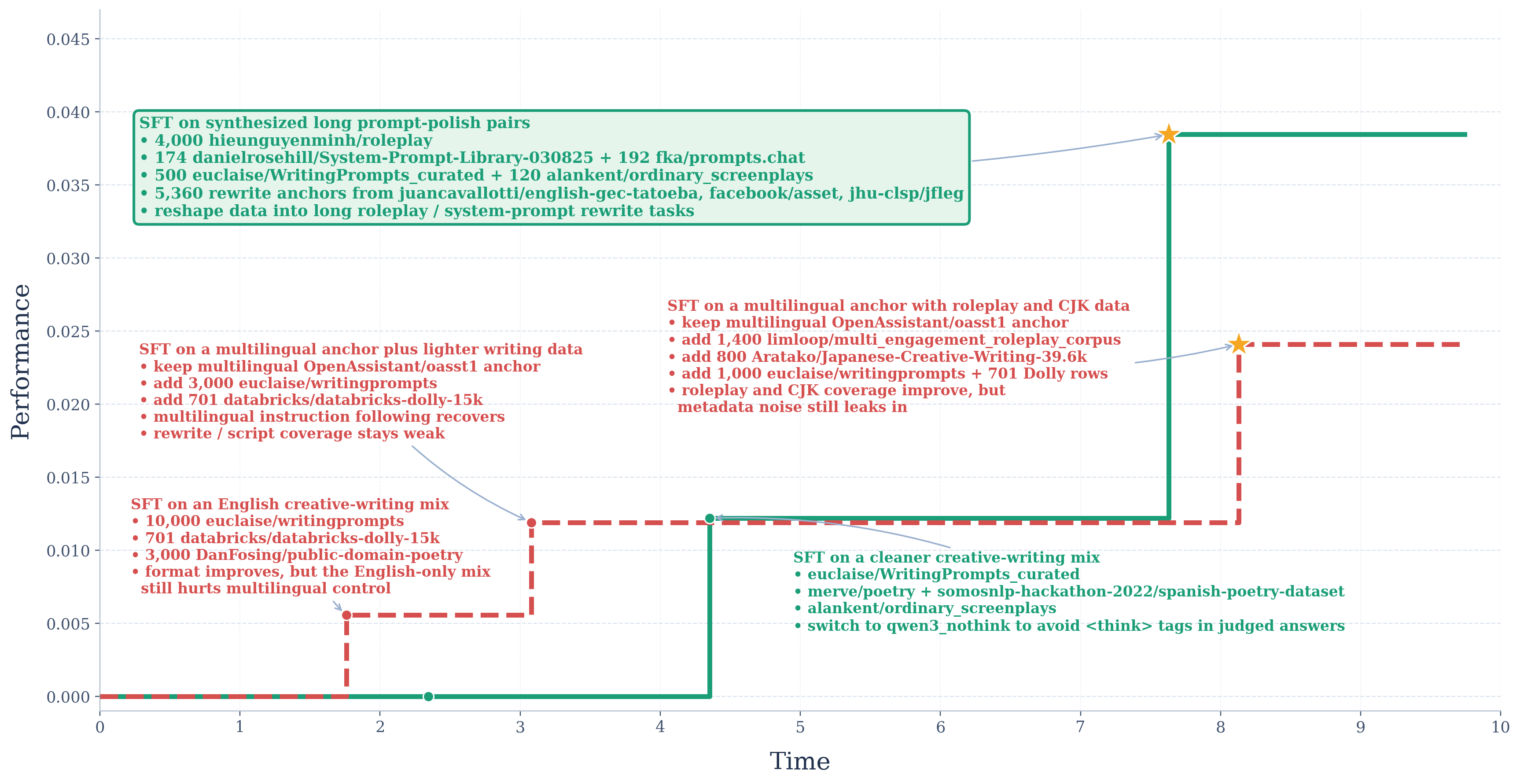}
    \caption{Ablation study of data skill on ArenaHard and Qwen3-4B.}
    \label{fig:arenahard-main-vs-data-ablation}
\end{figure}

\paragraph{Data skill helps the agent align training data with benchmark format and distribution.}
Figure~\ref{fig:arenahard-main-vs-data-ablation} shows a clear difference between the full system and the variant without the data skill. With the data skill, AutoTrainess eventually constructs a training mix centered on synthesized long prompt-polish pairs, rewrite anchors, and long roleplay or system-prompt rewrite tasks, which better match the style and distribution of ArenaHard. This route yields a substantially stronger final score. By contrast, the ablation version keeps trying broader or noisier writing mixtures, such as multilingual anchors plus lighter writing data or multilingual roleplay and CJK data, but fails to reshape them into the benchmark-facing supervision format required by the task. As a result, the no-data-skill variant improves only modestly and plateaus at a much lower level. This example suggests that the main benefit of the data skill is not simply giving the agent more data, but teaching it convert available data into a form whose format and distribution are much closer to the benchmark.

\begin{figure}[h]
    \centering
    \includegraphics[width=1\linewidth]{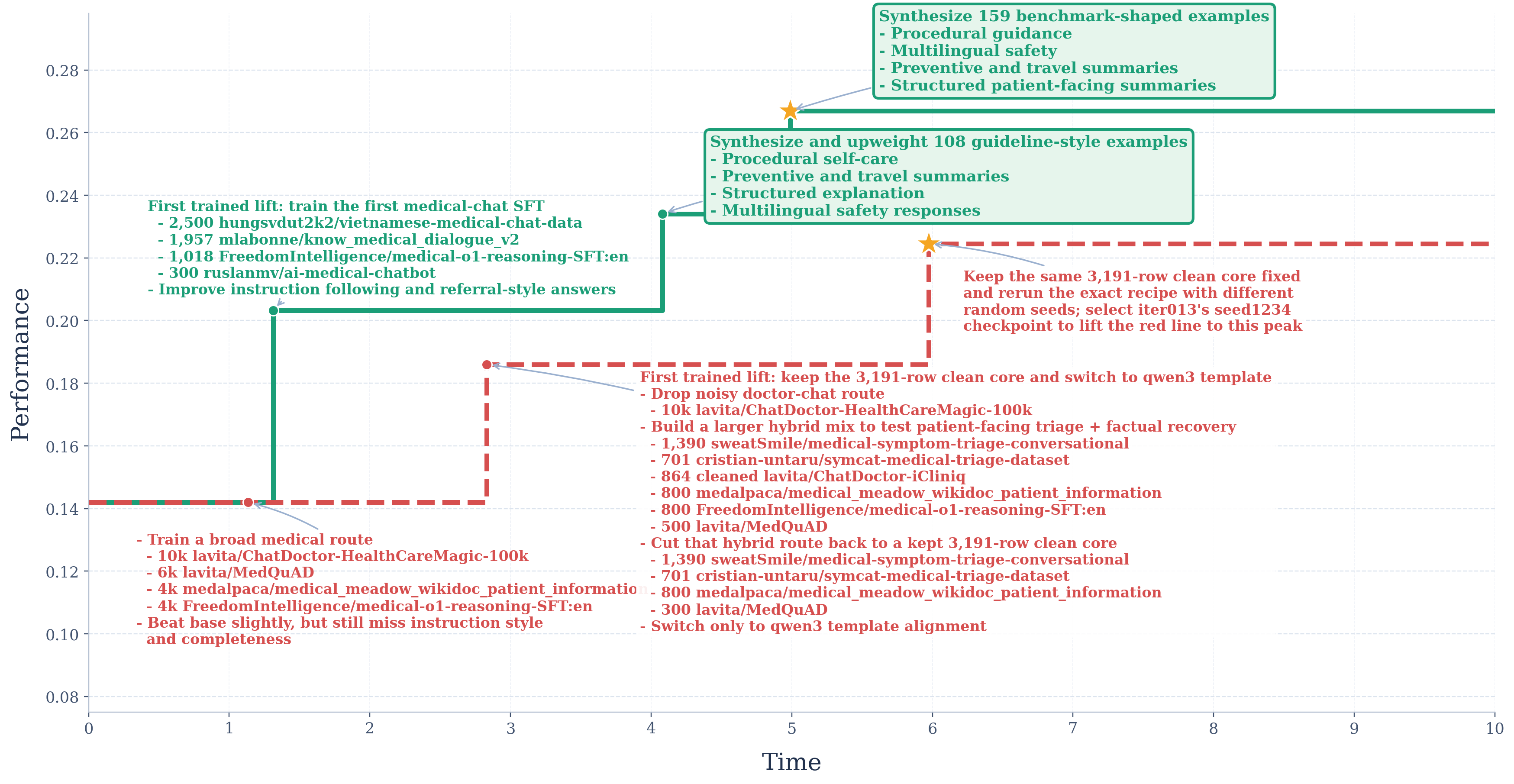}
    \caption{Ablation study of eval skill on HealthBench and Qwen3-4B.}
    \label{fig:arenahard-main-vs-eval-ablation}
\end{figure}

\paragraph{Eval skill helps the agent use evaluation evidence to drive targeted optimization.}
Figure~\ref{fig:arenahard-main-vs-eval-ablation} highlights a different failure mode. With the eval skill, AutoTrainess first improves the core medical-chat SFT mix, then uses evaluation feedback to synthesize and upweight benchmark-shaped examples that specifically target the remaining weaknesses, such as procedural guidance, preventive and travel summaries, multilingual safety, and structured patient-facing responses. This evaluation-guided shift produces another substantial gain late in the run. In contrast, the no-eval-skill ablation does not develop such targeted follow-up data. After an initial lift, it mostly keeps the same clean core fixed and continues by retrying different training parameters such as random seeds and different recipes, rather than constructing new supervision that addresses the concrete problems surfaced by evaluation. Consequently, its later-stage progress is much more limited. This case study suggests that the eval skill is valuable not only for measuring performance, but also for turning evaluation outcomes into actionable diagnoses that guide the next round of optimization.

\section{Detailed Explanation of the Agent Behavior Taxonomy}

In this section, we briefly explain the meaning of the behavior taxonomy used in Section~3.5.

\subsection{Evaluation Strategy}

\paragraph{$E1$ Subset Eval.}
In this iteration, the agent actually runs a sampled subset for validation. This is used for quick comparison, smoke testing, or low-cost iteration.

\paragraph{$E2$ Full Eval.}
In this iteration, the agent actually runs the full benchmark or full official evaluation set. 

\paragraph{$E3$ Fine-grained Eval.}
In this iteration, beyond evaluating the final result score, the agent conducts additional statistical analyses from multiple perspectives.

\subsection{Input Format Strategy}

\paragraph{$T1$ Prompt Alignment.}
The agent explicitly aligns the prompt to the benchmark prompt contract or user-message style in this iteration.

\paragraph{$T2$ Template Adjustment.}
The agent explicitly changes the model-side template or wrapper in this iteration. This includes chat-template changes, role/message schema changes, or think/non-think template switches.

\paragraph{$T3$ Prompt Style Change.}
The agent explicitly changes prompt wording, system instructions, or answer-behavior instructions without changing the core task family.

\subsection{Output Format Strategy}

\paragraph{$O1$ Direct Answer.}
In this iteration, the agent proposes training the model on data containing direct answers only, without explicit reasoning traces.

\paragraph{$O2$ Rationale + Answer.}
In this iteration, the agent proposes training the model on data that includes both rationales and final answers.

\paragraph{$O3$ Targeted Format.}
In this iteration, the agent proposes adopting a benchmark-facing output contract or an exact structured answer schema, such as `ANSWER: <number>` or `ANSWER: <LETTER>`.

\subsection{Data Strategy}

\paragraph{$D1$ Benchmark-related Data Source Selection.}
The agent chooses a data source that is highly similar to the benchmark task distribution. This is often an early move to find the most benchmark-like public data on huggingface.

\paragraph{$D2$ Training Dataset Expansion.}
The agent adds more data from the similar task family primarily to increase volume. The emphasis here is scale rather than special difficulty or narrow targeting.

\paragraph{$D3$ Data Difficulty Improvement.}
The agent deliberately chooses harder, expert-level, or challenge-style data. The main change is increasing task difficulty rather than simply increasing data size.

\paragraph{$D4$ Data Synthesis.}
The agent introduces genuinely new examples that were created rather than taken directly from an existing public dataset. This includes self-generated or programmatically constructed samples.

\paragraph{$D5$ Data Augmentation.}
The agent rewrites or transforms existing examples into new supervision views while keeping them grounded in the same source rows. The main change is creating derived variants rather than introducing a new source.

\paragraph{$D6$ Data Cleanup/Dedup/Rebalance.}
The agent cleans, deduplicates, trims, filters, or rebalances the dataset without introducing new data.

\subsection{Training Strategy}

\paragraph{$U1$ Full SFT.}
This iteration runs full-parameter supervised fine-tuning. The core update is standard full-model SFT.

\paragraph{$U2$ PEFT.}
This iteration runs parameter-efficient supervised fine-tuning. Typical examples include LoRA-style updates.

\paragraph{$U3$ DPO-style Training.}
This iteration runs preference-style training such as DPO, ORPO, or other chosen-versus-rejected objectives. 

\paragraph{$U4$ Self-distillation.}
This iteration runs distillation-style updates built from accepted or model-generated data.

\paragraph{$U5$ Continual Training.}
The training run in this iteration starts from a previously trained checkpoint.

\paragraph{$U6$ Train from Base Model.}
The training run in this iteration starts from the original base model. 

\paragraph{$U7$ Annealing Training.}
The agent uses a deliberately gentler continuation strategy, such as fewer steps or a lower learning rate. The main goal is to reduce update strength while continuing training.

\subsection{Planning Strategy}

\paragraph{$P1$ Baseline-based Planning.}
The agent runs a baseline first and uses the observed starting behavior to decide the first optimization direction. 

\paragraph{$P2$ Error Pattern Diagnosis.}
The agent diagnoses broad, recurring error patterns from observed failures, such as using the wrong template, producing truncated outputs, or repeatedly violating format constraints.

\paragraph{$P3$ Hypothesis Testing.}
The agent states or implies a concrete hypothesis for why the next change should help. The subsequent iteration is used to test whether the evidence supports that hypothesis.

\paragraph{$P4$ Failure Case Diagnosis.}
The agent focuses on specific remaining failure cases to identify the underlying capability gaps or unresolved weaknesses of the model. Rather than addressing broad error patterns, the agent examines concrete failed samples to determine which abilities are still insufficient.

\section{Instructions of AutoTrain Hub Workflow}

\subsection{AGENTS.md}
{
\begin{fileviewerbox}{AGENTS.md}
\begin{Verbatim}[breaklines=true,breakanywhere=true,fontsize=\small]
# AGENTS.md

## Objective
- Improve the target task or benchmark through real training and evaluation.
- Base important decisions on verifiable evidence.

## Hard Constraints
- APIs may be used only for benchmark evaluation, never for data construction or any other non-evaluation purpose.
- Do not use benchmark examples, the values of any fields or components of benchmark examples, or any data whose overlap with the benchmark cannot be ruled out for training or to create training data.
- Fine-tune only from the exact target base model provided by the task or checkpoints you fine-tuned from it; do not use an instruction-tuned, chat-tuned, larger, or different model as the training start point, merge source, fallback, or final submission.
- Never tune, search, or modify generation config to improve benchmark scores.
- If `CUDA_VISIBLE_DEVICES` is set, it defines the only GPU devices assigned to this run; do not override it or use any GPU outside that set.
- A run being blocked by environment, permissions, dependencies, shared resources, or turn-level execution limits does not mean the task is complete.

## Experiment Guidance
- Prefer the simplest valid approach first.
- Record the concrete reason for each failure, improvement, and strategy change.
- State which stage you are currently in.
- Do not spend iterations repeatedly tuning low-level settings unless there is clear evidence that doing so targets the current bottleneck.

## Stage Rules
- Stages are sequential and mandatory.
- If the current stage fails because of an engineering or environment issue, fix it and retry that stage.
- If the current stage has been explored enough to justify moving on, enter the next stage instead of ending the task.
- If no explicit target is provided, do not decide on your own that the task is complete.

## Iteration Execution
- For Stage 2/3, each stage is executed through one or more full iterations.
- Skills define how to execute one iteration locally. If a skill's default options conflict with the current stage rules, follow the stage rules in this file.
- A full iteration should use the skill chain: `iteration_plan -> data -> train -> eval -> log`.
- `iteration_plan` defines the goal and planned changes for the current iteration.
- `data` handles data selection, construction, and validation before training.
- `train` runs training and exports `final_model/`.
- `eval` runs the benchmark's real evaluation on `final_model/`.
- `log` records the completed iteration and its outcome.

## Procedure

### Stage 0: Task Definition
Define the target, evaluation method, and active resource constraints before training begins.
Once they are clear enough to support fair evaluation, proceed to Stage 1.

### Stage 1: Base Model Evaluation
Run the real benchmark evaluation on the current base model to establish the baseline result for later iterations.

Steps:
1. Prepare the current base model for evaluation.
2. Run the real benchmark evaluation.
3. Record the evaluation setup and result.

Decision rules:
- If an explicit target exists and the base model already reaches the target, stop.
- If evaluation fails because of an engineering or environment issue, fix the issue and repeat Stage 1. Otherwise, enter Stage 2.

### Stage 2: Local Diagnosis and Optimization
Run local iterations to establish a reliable training/evaluation loop, identify concrete failure modes, and test simple improvements before expanding the search space.

Iteration rules:
- Each iteration must target a specific observed weakness or opportunity from the latest valid evaluation.
- Prefer simple local changes with an explicit hypothesis.
- If an explicit target exists and the result reaches the target, stop.
- Continue Stage 2 while local iterations produce meaningful improvements, reveal new actionable evidence, or leave clear local hypotheses to test.
- Enter Stage 3 when local optimization has stalled, no clear local hypothesis remains, or the next likely improvement requires external methods, datasets, or training ideas.

### Stage 3: Evidence-Guided Exploration
Expand the search space beyond local optimization by using Stage 2 evidence to guide external search for methods, datasets, and training strategies.

Exploration routes:
- Problem-driven routes: target concrete failure modes, bottlenecks, or open hypotheses discovered in Stage 2.
- Open-ended routes: explore broader methods, datasets, or training ideas that may improve the benchmark even if they are not tied to one specific Stage 2 failure.

Iteration rules:
- Prioritize problem-driven routes when Stage 2 provides clear actionable evidence.
- Use open-ended routes when Stage 2 evidence is weak, local optimization has stalled, or broader benchmark knowledge suggests a promising direction.
- Each selected route must be tested through a full local iteration with an explicit hypothesis.
- If a route fails, record the reason and continue with the next concrete Stage 3 route unless a stop condition is met.

## Logging
- After each completed iteration, call the `log` skill to append the latest evidence to `task/experiment_log.md`.

## File Organization
- Cross-iteration summaries and decisions belong in `task/experiment_log.md`.
- Per-iteration files belong under `task/iterations/<iteration-id>/`, using the same iteration id as the experiment log.
- The iteration plan belongs in `task/iterations/<iteration-id>/iteration_plan.md`.
- Training data, data manifests, filtering notes, and validation notes for the iteration belong under `task/iterations/<iteration-id>/data/`.
- Training configs, commands, helper scripts, and training notes for the iteration belong under `task/iterations/<iteration-id>/train/`.
- Evaluation outputs, metrics, error traces, and evaluation notes for the iteration belong under `task/iterations/<iteration-id>/eval/`.
- Large checkpoints, copied datasets, caches, and temporary artifacts may be used when necessary, but their paths should be recorded and unnecessary copies should be removed promptly.
- The root `final_model/` directory is reserved for the current final submission model.

## Stop Conditions
- If an explicit target exists, the target is reached. Otherwise, no stop condition is available to the agent.

## Completion Criteria
- The task is complete only if a stop condition is met. Otherwise, you must not stop for any reason.
\end{Verbatim}
\end{fileviewerbox}
\captionsetup{type=figure}
\captionof{figure}{
\texttt{AGENTS.md} for AutoTrainess framework.
}
\label{fig.prompts.instance}
}

\subsection{Plan}
{
\begin{fileviewerbox}{Plan}
\begin{Verbatim}[breaklines=true,breakanywhere=true,fontsize=\small]
---
name: iteration_plan
description: Use when defining the goal and action plan for the next experiment iteration.
metadata:
  short-description: Plan the next experiment iteration
---

# iteration_plan

## Purpose
Define a clear goal and concrete action plan for the current experiment iteration based on real evidence from previous experiments.

## Inputs
- Results from previous experiments.
- Prior evaluation evidence and analysis.
- The current training and data context available in the workspace.

## Required outputs
- The main problems observed in previous experiments.
- The main objective of the current iteration.
- The changes planned for the current iteration.
- Whether this iteration mainly changes data, training, or both.
- The outcome that will count as success.
- Concise guidance for downstream data or training work.

## Rules
- Base the plan on real evidence from previous experiments rather than speculation.
- Focus on the main objective of the current iteration rather than trying to address every issue at once.
- Separate previous problems, current objective, planned changes, and success criteria clearly.
- Define the direction for the current iteration, but do not directly execute data construction or training in this skill.
- Keep the plan concrete enough that downstream skills can act on it.

## Procedure
1. Review previous experiment results and identify the main problems.
2. Decide what the current iteration is mainly trying to improve.
3. Define the main changes to make in this iteration.
4. State what outcome will count as success for this iteration.
5. Provide concise guidance for downstream data and training work.
\end{Verbatim}
\end{fileviewerbox}
\captionsetup{type=figure}
\captionof{figure}{
Instruction of \texttt{Plan} skill.
}
\label{fig.prompts.instance}
}

\subsection{Data Process}
{
\begin{fileviewerbox}{Data Process}
\begin{Verbatim}[breaklines=true,breakanywhere=true,fontsize=\small]
---
name: data
description: Use when preparing training data.
metadata:
  short-description: Prepare training data
---

# data

## Purpose
Prepare training data that addresses real problems exposed by previous training or evaluation, aligns with the benchmark evaluation interface, and is ready for downstream training.

## Core principles
- Drive all data work from concrete problems found in previous training or evaluation.
- Prioritize alignment with the benchmark's real evaluation interface over broad or generic data expansion.
- Prefer the smallest effective dataset that addresses the current problems.
- Keep data sources, transformations, and synthetic generation traceable.
- Avoid benchmark leakage and contaminated data.

## Workflow
1. Read [shared/conventions.md](./shared/conventions.md) for shared rules.
2. Run [selection/stage.md](./selection/stage.md) to identify target data needs and initial source directions.
3. Run [construction/stage.md](./construction/stage.md) to turn those needs and directions into a benchmark-aligned training dataset.
4. Run [validation/stage.md](./validation/stage.md) for data validation before training.
5. If validation finds construction issues, return to construction. If validation finds target-need or source-direction issues, return to selection.

## Required outputs
- A final training dataset ready for downstream training.
- A concise dataset description covering target problems, data sources, sample format, known limitations, and validation status.
\end{Verbatim}
\end{fileviewerbox}
\captionsetup{type=figure}
\captionof{figure}{
Instruction of \texttt{Data Process} skill.
}
\label{fig.prompts.instance}
}

\subsubsection{Selection}
{
\begin{fileviewerbox}{Data Process $\rightarrow$ Selection}
\begin{Verbatim}[breaklines=true,breakanywhere=true,fontsize=\small]
# Data Selection

## Purpose
Identify the data needs suggested by observed problems and choose initial source directions for construction.

## Required outputs
- The target problems or behaviors the new data should support.
- Initial source directions for construction.
- Important constraints or risks for construction, such as benchmark alignment, leakage risk, or source limitations.

## Rules
- Let observed problems guide the data direction.
- Avoid source directions that are clearly misaligned with the benchmark, low quality, or likely to introduce leakage or contamination.

## Procedure
1. Review available evidence from prior training, evaluation, or benchmark misses.
2. Identify the data needs implied by those problems or required benchmark-facing behaviors.
3. Choose initial source directions, such as local data, external data, synthetic data, or model-distilled data. If local or external data is substantially different from the benchmark distribution, consider synthetic or model-distilled data as source directions.
4. Pass unresolved assumptions, source limitations, leakage risks, and construction constraints to the construction stage.

\end{Verbatim}
\end{fileviewerbox}
\captionsetup{type=figure}
\captionof{figure}{
Instruction of \texttt{Selection} in \texttt{Data Process} skill.
}
\label{fig.prompts.instance}
}

\subsubsection{Construction}
{
\begin{fileviewerbox}{AGENTS}
\begin{Verbatim}[breaklines=true,breakanywhere=true,fontsize=\small]
# Data Construction

## Purpose
Turn the selected data needs and initial source directions into a benchmark-aligned training dataset.

## Required outputs
- A training dataset that is ready for downstream training.
- A concise dataset description covering source origins, transformations, target problems or behaviors, sample format, known limitations, and construction decisions.

## Rules
- Build data that supports the target problems or benchmark-facing behaviors, not generic capability expansion.
- Use the benchmark's `evaluate.py`, `templates/`, `task_context/`, or similar files when present to inspect the exact model-facing input and expected output form before deciding the training sample format.
- Align the training data with the benchmark's task interface, sample structure, input-output behavior, answer boundary, and final-answer location.
- Treat source directions from selection as starting points; return to selection if they cannot support the target data needs.
- Do not use any remote API or externally hosted model service for data construction; only local models are allowed.
- When local or externally collected data differs substantially from the benchmark distribution, consider bounded, explainable synthetic or model-distilled data that stays tied to the target data needs.
- Keep the dataset large enough to cover the target problems or behaviors, but small enough to avoid unnecessary noise, redundancy, and training cost.

## Construction actions
The agent may perform the following actions as needed:
- extract usable samples from source data
- clean corrupted or noisy data
- remove exact or obvious low-value duplicates
- reduce redundant or weakly relevant samples to keep the dataset focused
- rewrite or restructure samples to match the benchmark-facing format
- synthesize new samples
- distill new samples from local model outputs
- normalize fields into a consistent training schema

## Procedure
1. Review the target problems, source directions, constraints, and risks passed from selection.
2. Inspect the benchmark evaluation path and render or reconstruct several evaluation-style examples when possible.
3. Decide the target training sample format from the observed model-facing input, expected output form, answer boundary, and final-answer location.
4. Inspect candidate sources and decide whether they can support the target data needs. If they are viable, continue construction; if not, return to selection.
5. Extract, clean, rewrite, restructure, synthesize, or distill samples as needed.
6. Filter out broken, unreadable, empty, duplicated, misaligned, or clearly low-value samples, then reduce redundant or weakly relevant samples to keep the dataset focused.
7. Produce the final dataset and dataset description.

## Decision standard
The stage is complete when the dataset is usable for training, aligned with the benchmark-facing task, and described well enough for validation.
\end{Verbatim}
\end{fileviewerbox}
\captionsetup{type=figure}
\captionof{figure}{
Instruction of \texttt{Construction} in \texttt{Data Process} skill.
}
\label{fig.prompts.instance}
}

\subsubsection{Validation}
{
\begin{fileviewerbox}{Data Process $\rightarrow$ Validation}
\begin{Verbatim}[breaklines=true,breakanywhere=true,fontsize=\small]
# Data Validation

## Purpose
Validate the constructed dataset and dataset description before training, then decide whether the data is ready, needs reconstruction, or requires a new selection decision.

## Required outputs
- A validation result: approve for training, return to construction, or return to selection.
- A clear description of any detected problems and the reason for the return target.

## Validation dimensions

### 1. Structural validity
Check whether the data is mechanically usable:
- schema or template correctness
- field completeness and consistency
- readable encoding
- no broken, empty, truncated, or obviously malformed samples

### 2. Benchmark alignment
Check whether the data is training the right thing:
- input form matches the benchmark-facing task
- output form matches the benchmark-facing task
- constructed samples match the rendered evaluation-style examples
- answer boundaries and final-answer location match the rendered evaluation-style examples
- sample structure is compatible with the evaluation interface
- target behavior supports the target problems or behaviors described for the dataset

### 3. Content quality
Check whether the data itself is trustworthy:
- no obvious garbage text or unreadable content
- no clear sample errors
- no obvious low-value duplication
- no synthetic samples that are overly templated or unrealistic
- no leakage or contamination risk

## Rules
- Do not approve data only because the format looks correct; verify that the task objective is aligned as well.
- Do not approve data if sampled training examples do not match the rendered evaluation-style examples in input style, output style, answer boundary, and final-answer location.
- Separate construction errors from selection-direction errors.
- Return to construction if the direction is viable but the produced samples are flawed.
- Return to selection if the target data needs or initial source directions are not viable.

## Procedure
1. Inspect the constructed dataset and dataset description.
2. Check structural correctness, including schema, required fields, encoding, and malformed samples.
3. Compare several constructed training samples against the rendered evaluation-style examples.
4. Check whether the dataset matches the benchmark evaluation interface and target behaviors.
5. Review sample quality and look for garbage, corruption, duplication, leakage risk, or unrealistic synthesis.
6. Decide whether any detected problem belongs to construction or selection.
7. Produce one of three decisions:
   - approve for training
   - return to construction
   - return to selection

## Decision standard
The stage is complete when the dataset is approved for training or sent back with a clear reason and return target.
\end{Verbatim}
\end{fileviewerbox}
\captionsetup{type=figure}
\captionof{figure}{
Instruction of \texttt{Validation} in \texttt{Data Process} skill.
}
\label{fig.prompts.instance}
}

\subsection{Training}
{
\begin{fileviewerbox}{Training}
\begin{Verbatim}[breaklines=true,breakanywhere=true,fontsize=\small]
---
name: train
description: Use when running benchmark-oriented training with LlamaFactory, including SFT and RL.
metadata:
  short-description: Run training with LlamaFactory
---

# train

## Purpose
Run the simplest valid benchmark-oriented training workflow with LlamaFactory, choose the training mode that matches the current stage and evidence, and export a model ready for evaluation.

## When to use
- When training data is ready and the next step is to run model training.
- When the current stage requires supervised fine-tuning.
- When current evidence supports reinforcement learning.

## Core rules
- Use `hiyouga/LlamaFactory` for all training work.
- Read [shared/llamafactory.md](./shared/llamafactory.md) before running training.
- Choose the training mode that matches the current stage and evidence.
- Keep the workflow minimal and reproducible.
- Export `final_model/` for downstream evaluation.
- Do not switch to another framework or a custom training loop.

## Run caution
Long training jobs are allowed, but should be started deliberately. Prefer a short validation run first unless there is already clear evidence that a long run is necessary.

## Workflow
1. Read [shared/llamafactory.md](./shared/llamafactory.md).
2. Decide whether the current stage requires [sft/stage.md](./sft/stage.md) or [rl/stage.md](./rl/stage.md).
3. Follow the selected stage document.
4. Run training through the provided script in `scripts/`.
5. Export `final_model/` for evaluation.
\end{Verbatim}
\end{fileviewerbox}
\captionsetup{type=figure}
\captionof{figure}{
Instruction of \texttt{Training} skill.
}
\label{fig.prompts.instance}
}

\subsubsection{SFT}
{
\begin{fileviewerbox}{Training $\rightarrow$ SFT}
\begin{Verbatim}[breaklines=true,breakanywhere=true,fontsize=\small]
# SFT Stage

## Purpose
Run the minimum valid supervised fine-tuning workflow for the current stage with LlamaFactory.

## Inputs
- The training dataset prepared by the data workflow.
- The benchmark-facing sample format or schema.
- A valid base model path or model id.
- Any current-stage training settings required by the workspace.

## Rules
- Use `scripts/run_llamafactory.sh`.
- Use full-parameter fine-tuning only.
- Keep the training data aligned with the benchmark-facing task interface.
- Start with a small validation run before scaling up.
- Do not treat training loss alone as evidence of success.
- Keep the workflow minimal for the current stage.

## Procedure
1. Review the prepared training data and its benchmark-facing format.
2. Read `shared/llamafactory.md` and confirm that LlamaFactory is usable.
3. Prepare the minimum SFT dataset assets and verify the LlamaFactory config using `shared/llamafactory.md`.
4. Run a small validation training with `scripts/run_llamafactory.sh`.
5. If the validation run is usable, continue the intended SFT run.
6. Export `final_model/` and leave it ready for evaluation.

## Decision standard
The stage is complete only when the SFT run is reproducible, the exported model is evaluation-ready, and the result is not justified by training loss alone.
\end{Verbatim}
\end{fileviewerbox}
\captionsetup{type=figure}
\captionof{figure}{
Instruction of \texttt{SFT} in \texttt{Training} skill.
}
\label{fig.prompts.instance}
}

\subsubsection{RL}
{
\begin{fileviewerbox}{Training $\rightarrow$ RL}
\begin{Verbatim}[breaklines=true,breakanywhere=true,fontsize=\small]
# RL Stage

## Purpose
Run the minimum valid RL workflow for the current stage with LlamaFactory only when current evidence supports using RL.

## Inputs
- Recent evaluation evidence showing why RL is needed.
- The current model or base model to continue from.
- The minimum reward definition, feedback signal, or RL data required by the selected setup.
- Any current-stage constraints required by the workspace.

## Rules
- Use RL only when current evidence supports it.
- Use `scripts/run_llamafactory.sh`.
- State the reward or feedback signal actually used in practice.
- Start with a small validation run before scaling up.
- Stay inside the LlamaFactory workflow if failures occur.

## Procedure
1. Review the latest evaluation evidence and confirm that RL is justified.
2. Read `shared/llamafactory.md` and confirm that LlamaFactory is usable.
3. Prepare the minimum reward setup or RL data, and verify the LlamaFactory config using `shared/llamafactory.md`.
4. Run a small validation RL run with `scripts/run_llamafactory.sh`.
5. If the validation run is usable, continue the intended RL run.
6. Export `final_model/` and leave it ready for evaluation.

## Decision standard
The stage is complete only when RL is justified by current evidence, the run is reproducible, and the exported model is ready for real evaluation.
\end{Verbatim}
\end{fileviewerbox}
\captionsetup{type=figure}
\captionof{figure}{
Instruction of \texttt{RL} in \texttt{Training} skill.
}
\label{fig.prompts.instance}
}

\subsubsection{Shared Instruction}
{
\begin{fileviewerbox}{Training $\rightarrow$ Shared}
\begin{Verbatim}[breaklines=true,breakanywhere=true,fontsize=\small]
# LlamaFactory Workflow

## Purpose
Define the shared LlamaFactory workflow for training: installation, environment checks, execution boundary, and failure handling.

## Shared rules
- Use `hiyouga/LlamaFactory` for all training work.
- Do not replace it with another training framework or a custom training loop.
- If `LlamaFactory` is missing or broken, run `scripts/install_llamafactory.sh` first.
- When training fails, stay inside the LlamaFactory workflow and debug the real prerequisite, config, data, or resource issue.
- Leave the final result ready for downstream evaluation.

## Prerequisite checks
Before running training, confirm the following:
- `LlamaFactory` is installed and runnable.
- The relevant script under `scripts/` is present.
- A valid base model path or model id is available.
- The current Python environment and required dependencies are usable.
- The required training inputs for the selected mode are available.

## Config discipline
Before running LlamaFactory, inspect the training config instead of relying on defaults or memory.

Pay special attention to parameters that strongly affect training behavior:
- `model_name_or_path`: use the intended base model or checkpoint.
- `finetuning_type`: keep it consistent with the selected stage; do not switch to LoRA when full-parameter fine-tuning is required.
- `template`: must match the benchmark-facing model input format used by the evaluation path; do not use a default or a random template without verifying this match.
- `dataset` and `dataset_dir`: point to the current prepared training data.
- `learning_rate`, `num_train_epochs` or `max_steps`, `per_device_train_batch_size`, and `gradient_accumulation_steps`: keep the training strength deliberate.
- `cutoff_len`: avoid truncating important prompt, reasoning, or answer content.

If an existing workspace config is available, start from it and change only what the current run requires.

## Failure handling
- If the failure indicates that `LlamaFactory` or its CLI is missing or broken, run `scripts/install_llamafactory.sh` first.
- If LlamaFactory fails with a dependency version check such as `transformers>=...,<=... is required`, use `DISABLE_VERSION_CHECK=1` for the retry. The provided `scripts/run_llamafactory.sh` already sets this for training runs.
- If the failure is caused by data, config, or resource issues, fix those issues without switching frameworks.
- After fixing a failure, rerun a small validation job before continuing the intended run.

## Output expectation
- Training should remain reproducible within the current workspace setup.
- The result should be exported as `final_model/`.
- `final_model/` must be loadable by `vllm`.
- The exported model should be left ready for real evaluation rather than as an unfinished intermediate artifact.
\end{Verbatim}
\end{fileviewerbox}
\captionsetup{type=figure}
\captionof{figure}{
Shared instruction in \texttt{Training} skill.
}
\label{fig.prompts.instance}
}

\subsection{Evaluation}
{
\begin{fileviewerbox}{Evaluation}
\begin{Verbatim}[breaklines=true,breakanywhere=true,fontsize=\small]
---
name: eval
description: Use when evaluating a model on the current benchmark(s).
metadata:
  short-description: Run benchmark evaluation
---

# eval

## Purpose
Run the benchmark's real evaluation on `final_model/` and record reproducible evidence needed for the next stage decision.

## Inputs
- Workspace repository (current working directory).
- `final_model/`.

## Required outputs
- `eval_results/` with raw outputs or logs.
- The exact evaluation command or config used.
- A concise metrics summary.
- `eval_results/sample_summary.md` with 15 randomly selected evaluation samples, including score, input, target, and model output.
- A brief note on the main 1-3 observed failure modes and whether each one looks more like a data problem, a training problem, or an inference/template problem.

## Rules
- Use the benchmark's real evaluation entrypoint.
- If evaluation fails, stay in the benchmark's real evaluation workflow, debug the failure, and retry.
- For any evaluation used to compare checkpoints, judge model quality, or choose the next iteration, use at least `max(32, ceil(5% of the benchmark))` samples. If the benchmark has fewer than 32 samples, evaluate the full benchmark.
- Runs below that sample floor are allowed only as smoke tests for command or runtime validity; do not use them as evidence that one checkpoint or approach is better.
- Always produce `eval_results/sample_summary.md` with 15 random evaluation samples.
- Use `skills/eval/scripts/summarize_eval_samples.py` when the benchmark outputs compatible `inspect_ai` logs; otherwise, add the minimum benchmark-specific script or logging needed to generate the sample summary from the real evaluation run.
- Keep the output focused on evidence needed for the next decision.

## Procedure
1) Locate the canonical evaluation entrypoint.
2) If using a limited evaluation, determine the benchmark sample count and choose a limit that satisfies the sample-floor rule.
3) Run evaluation on `final_model/`.
4) Save raw outputs, commands, the sample count or limit used, and a concise metrics summary under `eval_results/`.
5) If evaluation fails, debug it inside the benchmark's real evaluation workflow, then retry with the minimum necessary fix.
6) Generate `eval_results/sample_summary.md` with 15 random samples including score, input, target, and model output. Use `skills/eval/scripts/summarize_eval_samples.py` when compatible `inspect_ai` logs are available; otherwise, add the minimum benchmark-specific script or logging needed.
7) Verify that `eval_results/sample_summary.md` was generated and contains 15 samples.
8) Summarize the main 1-3 observed failure modes and whether each one looks more like a data problem, a training problem, or an inference/template problem.
\end{Verbatim}
\end{fileviewerbox}
\captionsetup{type=figure}
\captionof{figure}{
Instruction of \texttt{Evaluation} skill.
}
\label{fig.prompts.instance}
}

\subsection{Log}
{
\begin{fileviewerbox}{Log}
\begin{Verbatim}[breaklines=true,breakanywhere=true,fontsize=\small]
---
name: log
description: Use when appending an experiment log entry after a completed iteration.
metadata:
  short-description: Append experiment log
---

# log

## Task
After each completed iteration, append one new entry to `task/experiment_log.md`.

## Rules
- If `task/experiment_log.md` does not exist, create it. If it already exists, append a new entry at the end.
- Each call should record only the iteration that has just finished.
- Organize the log by stage, and record the work done in the current iteration under the relevant stage.

## Entry Format
Organize entries by stage when a stage is available. If the relevant stage heading does not exist, create it.

Use this Markdown format for each new entry:

### Iteration <id>: <short title>

- Context: <stage, objective, or current focus>
- Status: completed | failed | blocked
- Motivation: <what prompted this iteration>
- References: <papers, docs, repos, datasets, blogs, or notes consulted; write "None" if not used>
- Starting checkpoint: <base model, previous checkpoint, or final_model path used as training start>
- Training data: <datasets/files used, sizes, filters, construction method, validation notes>
- Method: <training method, recipe, prompt/data strategy, or implementation changes>
- Training config: <key hyperparameters, command, epochs, lr, batch size, LoRA/full fine-tune, etc.>
- Evaluation: <evaluation command, benchmark split, limit/full setting, metric>
- Result: <exact score, failure, or observed behavior>
- Analysis: <what changed, what likely caused it, whether the hypothesis was supported>
- Artifacts: <model path, logs, data files, checkpoints>
- Next action: <the justified next step>

Rules:
- Fill every field. Use `None` or `N/A` only when the field truly does not apply.
- Record concrete evidence, not vague summaries.
- Include exact metrics, commands, paths, and dataset sizes when available.
- If references were consulted, record enough detail to identify them later.
- If the iteration failed or was blocked, record the specific cause.
- The next action must follow from the recorded result and analysis.
\end{Verbatim}
\end{fileviewerbox}
\captionsetup{type=figure}
\captionof{figure}{
Instruction of \texttt{Log} skill.
}
\label{fig.prompts.instance}
}
\section{Detailed Results on PostTrainBench}
\label{app-detail-res}
\begin{table}[h]
\caption{Results of AIME2025 on PostTrainBench.}
\centering

\resizebox{\linewidth}{!}{%
\begin{tabular}{lccccc}
\toprule
 & \multicolumn{5}{c}{\textbf{AIME2025}} \\
\cmidrule(lr){2-6}
\textbf{Harness} & Qwen3-1.7B & Qwen3-4B & SmolLM-3B & Gemma-4B & \textbf{Avg.} \\
\cmidrule(lr){1-1} \cmidrule(lr){2-6}

Instruct     & 26.67 & 53.33 & 26.67 & 10.00 & 29.17 \\
Base    & 0.00  & 3.33 & 3.33  & 0.00  & 1.67 \\
\cmidrule(lr){1-1} 
CLI-only    &   &   &   &   &   \\
\smalltab w/ GPT-5.4 (Codex)   & 0.00 & 6.67 & 3.33 & 3.33 & 3.33 \\
\smalltab w/ GPT-5.4 (OpenCode)   & 3.33    & 0.00    & 0.00   & 0.00   & 0.83 \\
\smalltab w/ DeepSeek-V4-Flash (OpenCode) & 0.00    & 3.33    & 0.00   & 0.00   & 0.83 \\
\cmidrule(lr){1-1} 
AutoTrainess &     &     &     &     &  \\
\smalltab w/ GPT-5.4 (Codex) & 0.00 & 6.67 & 0.00 & 0.00 & 1.67 \\
\smalltab w/ GPT-5.4 (OpenCode)   & 0.00    & 0.00    & 0.00    & 0.00    & 0.00 \\
\smalltab w/ DeepSeek-V4-Flash (OpenCode) & 0.00    & 0.00    & 0.00    & 0.00    & 0.00 \\
\bottomrule
\end{tabular}
}
\end{table}
\begin{table}[h]
\caption{Results of ArenaHard Writing on PostTrainBench.}
\centering

\resizebox{\linewidth}{!}{%
\begin{tabular}{lccccc}
\toprule
 & \multicolumn{5}{c}{\textbf{ArenaHard Writing}} \\
\cmidrule(lr){2-6}
\textbf{Harness} & Qwen3-1.7B & Qwen3-4B & SmolLM-3B & Gemma-4B & \textbf{Avg.} \\
\cmidrule(lr){1-1} \cmidrule(lr){2-6}

Instruct     & 50.00 & 86.84 & 49.20 & 94.80 & 70.21 \\
Base    & 0.91  & 3.42 & 0.42 & 0.29  & 1.26 \\
\cmidrule(lr){1-1} 
CLI-only    &   &   &   &   &   \\
\smalltab w/ GPT-5.4 (Codex)   & 3.89 & 3.36 & 3.61 & 0.29 & 2.79 \\
\smalltab w/ GPT-5.4 (OpenCode)   & 0.59 & 1.50 & 0.21 & 5.37 & 1.92 \\
\smalltab w/ DeepSeek-V4-Flash (OpenCode) & 1.09 & 2.54 & 0.57 & 0.29 & 1.12 \\
\cmidrule(lr){1-1} 
AutoTrainess &     &     &     &     &  \\
\smalltab w/ GPT-5.4 (Codex) & 0.32 & 1.41 & 0.10 & 16.76 & 4.65 \\
\smalltab w/ GPT-5.4 (OpenCode)   & 0.50    & 4.67    & 1.36    & 3.59    & 2.53 \\
\smalltab w/ DeepSeek-V4-Flash (OpenCode) & 0.74   & 10.23    & 7.17    & 6.15    & 6.07 \\
\bottomrule
\end{tabular}
}
\end{table}
\begin{table}[h]
\caption{Results of BFCL on PostTrainBench.}
\centering

\resizebox{\linewidth}{!}{%
\begin{tabular}{lccccc}
\toprule
 & \multicolumn{5}{c}{\textbf{BFCL}} \\
\cmidrule(lr){2-6}
\textbf{Harness} & Qwen3-1.7B & Qwen3-4B & SmolLM-3B & Gemma-4B & \textbf{Avg.} \\
\cmidrule(lr){1-1} \cmidrule(lr){2-6}

Instruct     & 94.00 & 95.00 & 84.00 & 67.00 & 85.00 \\
Base    & 0.00  & 0.00 & 0.00 & 6.00  & 1.50 \\
\cmidrule(lr){1-1} 
CLI-only    &   &   &   &   &   \\
\smalltab w/ GPT-5.4 (Codex)   & 54.00 & 55.00 & 91.00 & 96.00 & 74.00 \\
\smalltab w/ GPT-5.4 (OpenCode)   & 98.00   & 0.00    & 0.00  & 97.00   & 48.75 \\
\smalltab w/ DeepSeek-V4-Flash (OpenCode) & 0.00    & 0.00    & 0.00   & 6.00   & 1.50 \\
\cmidrule(lr){1-1} 
AutoTrainess &     &     &     &     &  \\
\smalltab w/ GPT-5.4 (Codex) & 90.00 & 100.00 & 97.00 & 100.00 & 96.75 \\
\smalltab w/ GPT-5.4 (OpenCode)   & 90.00    & 95.00    & 95.00    & 92.00    & 93.00 \\
\smalltab w/ DeepSeek-V4-Flash (OpenCode) & 0.00   & 0.00    & 0.00    & 70.00    & 17.50 \\
\bottomrule
\end{tabular}
}
\end{table}
\begin{table}[h]
\caption{Results of GPQA Main on PostTrainBench.}
\centering

\resizebox{\linewidth}{!}{%
\begin{tabular}{lccccc}
\toprule
 & \multicolumn{5}{c}{\textbf{GPQA Main}} \\
\cmidrule(lr){2-6}
\textbf{Harness} & Qwen3-1.7B & Qwen3-4B & SmolLM-3B & Gemma-4B & \textbf{Avg.} \\
\cmidrule(lr){1-1} \cmidrule(lr){2-6}

Instruct     & 35.49 & 44.64 & 33.26 & 31.47 & 36.22 \\
Base    & 14.06  & 13.39 & 4.91  & 1.56  & 8.48 \\
\cmidrule(lr){1-1} 
CLI-only    &   &   &   &   &   \\
\smalltab w/ GPT-5.4 (Codex)   & 11.16 & 27.68 & 22.54 & 25.22 & 21.65 \\
\smalltab w/ GPT-5.4 (OpenCode)   & 25.67    & 25.00    & 26.12  & 24.33   & 25.28 \\
\smalltab w/ DeepSeek-V4-Flash (OpenCode) & 14.06    & 13.39    & 28.35   & 1.56   & 14.34 \\
\cmidrule(lr){1-1} 
AutoTrainess &     &     &     &     &  \\
\smalltab w/ GPT-5.4 (Codex) & 29.91 & 35.71 & 28.13 & 22.99 & 29.19 \\
\smalltab w/ GPT-5.4 (OpenCode)   & 29.46    & 32.81    & 26.34    & 23.44    & 28.01 \\
\smalltab w/ DeepSeek-V4-Flash (OpenCode) & 30.13   & 13.39    & 23.44    & 31.03    & 24.50 \\
\bottomrule
\end{tabular}
}
\end{table}
\begin{table}[h]
\caption{Results of GSM8K on PostTrainBench.}
\centering

\resizebox{\linewidth}{!}{%
\begin{tabular}{lccccc}
\toprule
 & \multicolumn{5}{c}{\textbf{GSM8K}} \\
\cmidrule(lr){2-6}
\textbf{Harness} & Qwen3-1.7B & Qwen3-4B & SmolLM-3B & Gemma-4B & \textbf{Avg.} \\
\cmidrule(lr){1-1} \cmidrule(lr){2-6}

Instruct     & 88.48 & 93.78 & 82.18 & 83.55 & 87.00 \\
Base    & 12.66  & 41.85 & 21.08  & 6.14  & 20.43 \\
\cmidrule(lr){1-1} 
CLI-only    &   &   &   &   &   \\
\smalltab w/ GPT-5.4 (Codex)   & 44.20 & 42.38 & 64.67 & 52.99 & 51.06 \\
\smalltab w/ GPT-5.4 (OpenCode)   & 25.93 & 41.85 & 70.36 & 31.69 & 42.46 \\
\smalltab w/ DeepSeek-V4-Flash (OpenCode) & 4.17  & 47.31  & 54.66 & 34.87 & 35.25 \\
\cmidrule(lr){1-1} 
AutoTrainess &     &     &     &     &  \\
\smalltab w/ GPT-5.4 (Codex) & 54.74 & 75.59 & 57.70 & 40.64 & 57.17 \\
\smalltab w/ GPT-5.4 (OpenCode)   & 54.44    & 45.94    & 50.34    & 14.25    & 41.24 \\
\smalltab w/ DeepSeek-V4-Flash (OpenCode) & 57.70   & 57.47 & 51.93  & 52.46  & 54.89 \\
\bottomrule
\end{tabular}
}
\end{table}
\begin{table}[h]
\caption{Results of HealthBench Easy on PostTrainBench.}
\centering

\resizebox{\linewidth}{!}{%
\begin{tabular}{lccccc}
\toprule
 & \multicolumn{5}{c}{\textbf{HealthBench Easy}} \\
\cmidrule(lr){2-6}
\textbf{Harness} & Qwen3-1.7B & Qwen3-4B & SmolLM-3B & Gemma-4B & \textbf{Avg.} \\
\cmidrule(lr){1-1} \cmidrule(lr){2-6}

Instruct     & 44.92 & 52.72 & 29.58 & 46.06 & 43.32 \\
Base    & 7.54  & 13.38 & 0.00  & 17.04  & 9.49 \\
\cmidrule(lr){1-1} 
CLI-only    &   &   &   &   &   \\
\smalltab w/ GPT-5.4 (Codex)   & 15.06 & 21.09 & 7.34 & 18.09 & 15.40 \\
\smalltab w/ GPT-5.4 (OpenCode)   & 14.09    & 18.80    & 15.38  & 14.13   & 15.60 \\
\smalltab w/ DeepSeek-V4-Flash (OpenCode) & 2.98    & 7.91    & 6.52   & 20.39   & 9.45 \\
\cmidrule(lr){1-1} 
AutoTrainess &     &     &     &     &  \\
\smalltab w/ GPT-5.4 (Codex) & 14.95 & 18.20 & 18.82 & 12.22 & 16.05 \\
\smalltab w/ GPT-5.4 (OpenCode)   & 6.94    & 13.38    & 18.37    & 17.04    & 13.93 \\
\smalltab w/ DeepSeek-V4-Flash (OpenCode) & 3.29   & 27.18  & 23.52  & 16.55  & 17.64 \\
\bottomrule
\end{tabular}
}
\end{table}
\begin{table}[h]
\caption{Results of HumanEval on PostTrainBench.}
\centering

\resizebox{\linewidth}{!}{%
\begin{tabular}{lccccc}
\toprule
 & \multicolumn{5}{c}{\textbf{HumanEval}} \\
\cmidrule(lr){2-6}
\textbf{Harness} & Qwen3-1.7B & Qwen3-4B & SmolLM-3B & Gemma-4B & \textbf{Avg.} \\
\cmidrule(lr){1-1} \cmidrule(lr){2-6}

Instruct     & 68.90 & 77.44 & 70.12 & 69.51 & 71.49 \\
Base    & 7.93  & 36.59 & 6.10  & 0.61  & 12.81 \\
\cmidrule(lr){1-1} 
CLI-only    &   &   &   &   &   \\
\smalltab w/ GPT-5.4 (Codex)   & 29.27 & 63.41 & 34.15 & 28.05 & 38.72 \\
\smalltab w/ GPT-5.4 (OpenCode)   & 10.37    & 36.59    & 39.63  & 33.54  & 30.03 \\
\smalltab w/ DeepSeek-V4-Flash (OpenCode) & 37.20   & 50.00  & 30.49  & 24.39  & 35.52 \\
\cmidrule(lr){1-1} 
AutoTrainess &     &     &     &     &  \\
\smalltab w/ GPT-5.4 (Codex) & 40.85 & 47.56 & 29.88 & 34.76 & 38.26 \\
\smalltab w/ GPT-5.4 (OpenCode)   & 21.95   & 40.24    & 28.05    & 40.24    & 32.62 \\
\smalltab w/ DeepSeek-V4-Flash (OpenCode) & 36.59  & 70.12  & 6.70   & 31.10  & 36.13 \\
\bottomrule
\end{tabular}
}
\end{table}




\newpage

\end{document}